\documentclass[journal]{IEEEtran}
\usepackage{tipa}
\usepackage{bbm}
\usepackage{mathrsfs}
\usepackage{cite,url,subfigure,epsfig,graphicx}
\usepackage{amssymb,amsmath}
\usepackage{indentfirst}
\usepackage{algorithmic}
\usepackage{algorithm}
\usepackage{pdfpages}
\usepackage{times,verbatim,amsfonts,amsmath,color}
\usepackage{pifont}
\usepackage{multirow}
\usepackage{booktabs}
\usepackage{adjustbox}
\usepackage{hyperref}
\usepackage{threeparttable}
\usepackage{pifont}

\usepackage{booktabs}  
\usepackage{array}    
\usepackage{makecell} 
\usepackage{tikz}
\usepackage[edges]{forest}

\IEEEoverridecommandlockouts
\makeatletter
\def\ps@headings{\def\@oddhead{\mbox{}\scriptsize\rightmark \hfil \thepage}\def\@evenhead{\scriptsize\thepage \hfil \leftmark\mbox{}}\def\@oddfoot{}\def\@evenfoot{}}
\makeatother 

\usepackage{geometry}
\begin{document}
\title{Security of OpenClaw Agents: Fundamentals, Threats, and Countermeasures}
\author{Yuntao~Wang, Jianle~Ba, Han~Liu, Yanghe~Pan, Jintao~Wei, Zhou~Su, Tom~H.~Luan, and Linkang~Du
\thanks{Y.~Wang, J.~Ba, H~Liu, Y.~Pan, J.~Wei, Z.~Su, T.~H.~Luan, and L.~Du are with the School of Cyber Science and Engineering, Xi'an Jiaotong University, Xi'an, China \textit{(Corresponding author: Zhou~Su).}}}

\maketitle

\begin{abstract}
The rapid evolution of large language model (LLM)-driven autonomous agents has given rise to OpenClaw, a new class of open-source agent frameworks that operate as continuously running, skill-augmented systems with persistent memory, multi-channel interaction, and high degrees of autonomy. 
Such capabilities enable OpenClaw agents to autonomously execute complex, multi-step tasks and interact seamlessly with external applications, but simultaneously introduce a substantially enlarged attack surface. In particular, the combination of high-privilege operations and persistent memory exposes OpenClaw agents to various emerging threats, including skill poisoning, cognitive manipulation, multi-agent cascading failures, and supply-chain vulnerabilities.
In this survey, we present a comprehensive study of the security landscape of OpenClaw agents. We first examine the general architecture and key characteristics that distinguish OpenClaw agents from traditional AI agent systems. 
We categorize existing security and privacy threats into a layered framework and analyze how vulnerabilities arise during agent reasoning, action execution, and external interaction. Representative defense mechanisms are also reviewed to draw the current defense landscape. Finally, several unresolved issues related to the reliability and trustworthiness of OpenClaw ecosystems are discussed.
\end{abstract}


\IEEEpeerreviewmaketitle
\section{Introduction}
Recent advances in large language models (LLMs) have enabled autonomous agents that can interact with environments, perform multi-step reasoning, and invoke external tools to accomplish tasks \cite{wang2026large}. Among them, OpenClaw represents a new generation of open-source autonomous agents that run on user-controlled devices or the cloud and interact with users through multiple communication channels such as messaging platforms and web interfaces \cite{OpenClawWebsite}. 

Built as a full agent runtime rather than a simple chatbot wrapper, OpenClaw integrates core components including multi-channel interaction gateways, persistent memory management, and modular skill orchestration \cite{li2026openclaw}. This design enables OpenClaw agents to operate as continuously running entities that autonomously decompose high-level objectives, coordinate multiple skills, interact with external applications, and adapt their behavior through iterative reasoning and contextual feedback. As a result, OpenClaw agents can execute complex multi-step workflows, ranging from automated document processing and form filling to online social engagement \cite{chen2026openclaw}. 
OpenClaw has attracted remarkable attention since its public release in November 2025 and has accumulated over 330k GitHub stars, making it one of the fastest-growing open-source frameworks \cite{OpenClawGithub}. The skill marketplace also allows users to install third-party skills developed by the community, greatly expanding the functionality of OpenClaw agents.

Unlike conventional LLM-based applications, OpenClaw agents can directly access local resources, invoke external skills, and interact with other applications or agents \cite{ying2026uncovering,chen2026trajectory}. Recent work has shown that such capabilities can introduce new attack surfaces, e.g., prompt injection attacks can alter task execution flows \cite{cheng2026agent}, malicious skills may trigger unintended operations \cite{dong2026clawdrain}, and memory poisoning can tamper with stored contexts or long-term agent states \cite{wang2026assistant}.
Specifically, as OpenClaw agents bridge LLM outputs with executable system operations, model-level failures may propagate to connected applications or host environments \cite{wang2026assistant}.
As reported by Oasis Security, a malicious webpage could hijack locally deployed OpenClaw agents during ordinary browsing tasks, even without browser plugins or explicit user interaction \cite{OpenClawAgentTakeover}. 
The OpenClaw skill ecosystem also raises supply-chain security risks. Cisco reported that 26\% of about 31k publicly available agent skills contained at least one vulnerability \cite{CiscoOpenClaw}. Such vulnerabilities may expose sensitive information stored on user devices, including API keys, chat histories, and financial documents.
Lastly, as OpenClaw agents become increasingly interconnected, vulnerabilities originating from a single agent may propagate across other components, agents, or services within the ecosystem \cite{zhang2026clawworm}. Therefore, defending OpenClaw agents requires security mechanisms that can jointly protect models, memory, skills, and external tool interactions \cite{liu2026clawkeeper}.


\subsection{Comparison with Existing Surveys and Our Contributions}\label{subsec:Contributions}

A growing number of survey papers have studied the security of AI agents. 
Deng \emph{et al.} \cite{deng2025ai} present a comprehensive survey on security threats of AI agents from four aspects: intra-execution vulnerabilities, memory-related risks, operational environment threats, and inter-agent interactions. 
He \emph{et al.} \cite{he2025emerged} comprehensively review AI agent security from two complementary perspectives: LLM-related vulnerabilities and agent-specific threats. They also analyze potential impacts on humans, environments, and other agents. 
Yu \emph{et al.} \cite{yu2025survey} further survey the trustworthiness-related threats of AI agent systems and classify them into intrinsic components (e.g., brain, tools, and memory) and extrinsic components (e.g., users, agents, and environments). 
Beyond standalone agents, Wang \emph{et al.} \cite{Wang2025IOA} investigate security risks in the Internet of Agents (IoA), including identity authentication, embodied AI security, cross-agent trust, and privacy protection. 

Existing surveys mainly discuss general-purpose LLM agents and multi-agent frameworks.
Compared with conventional LLM agents, OpenClaw systems integrate persistent memory, autonomous skill execution, and continuous task coordination, which introduces an additional attack surface.
Several recent surveys have explored the security of OpenClaw-style agents. Ying \emph{et al.} \cite{ying2026uncovering} design a layered security architecture of OpenClaw agents and categorize OpenClaw security issues into three layers: software and execution security, AI security, and information and system security. 
Deng \emph{et al.} \cite{deng2026taming} analyze OpenClaw security threats from a lifecycle perspective across five operational stages: initialization, input, inference, decision, and execution. They further summarize defense approaches associated with different phases of agent operation.

Different from prior surveys, this work focuses specifically on the security of OpenClaw agents and analyzes the risks introduced by their architectural design and operational workflow. Firstly, we analyze the architectural components and key characteristics of OpenClaw agents that give rise to unique security challenges. We then present a unified taxonomy of OpenClaw-specific threats and organize them across the cognition, execution, and interaction levels, followed by a discussion of representative defense strategies and emerging countermeasures.
The main contributions of this survey include:
\begin{itemize}
\item We analyze the architectural components of OpenClaw agents, including the agent runtime framework, persistent memory mechanisms, and skill orchestration process. We also discuss the unique characteristics of OpenClaw agents that differ from conventional AI agents.
\item We systematically identify and categorize the security and privacy threats specific to OpenClaw agents, covering vulnerabilities across model reasoning loop, skill ecosystem, memory management, and execution environments.
\item We review existing defense mechanisms and emerging countermeasures for securing OpenClaw agents, and further discuss open research challenges and future directions toward building trustworthy and resilient OpenClaw-based agent ecosystems.
\end{itemize}

\begin{figure}[!t]
\centering \setlength{\abovecaptionskip}{-0.1cm}
\includegraphics[width=8.76cm]{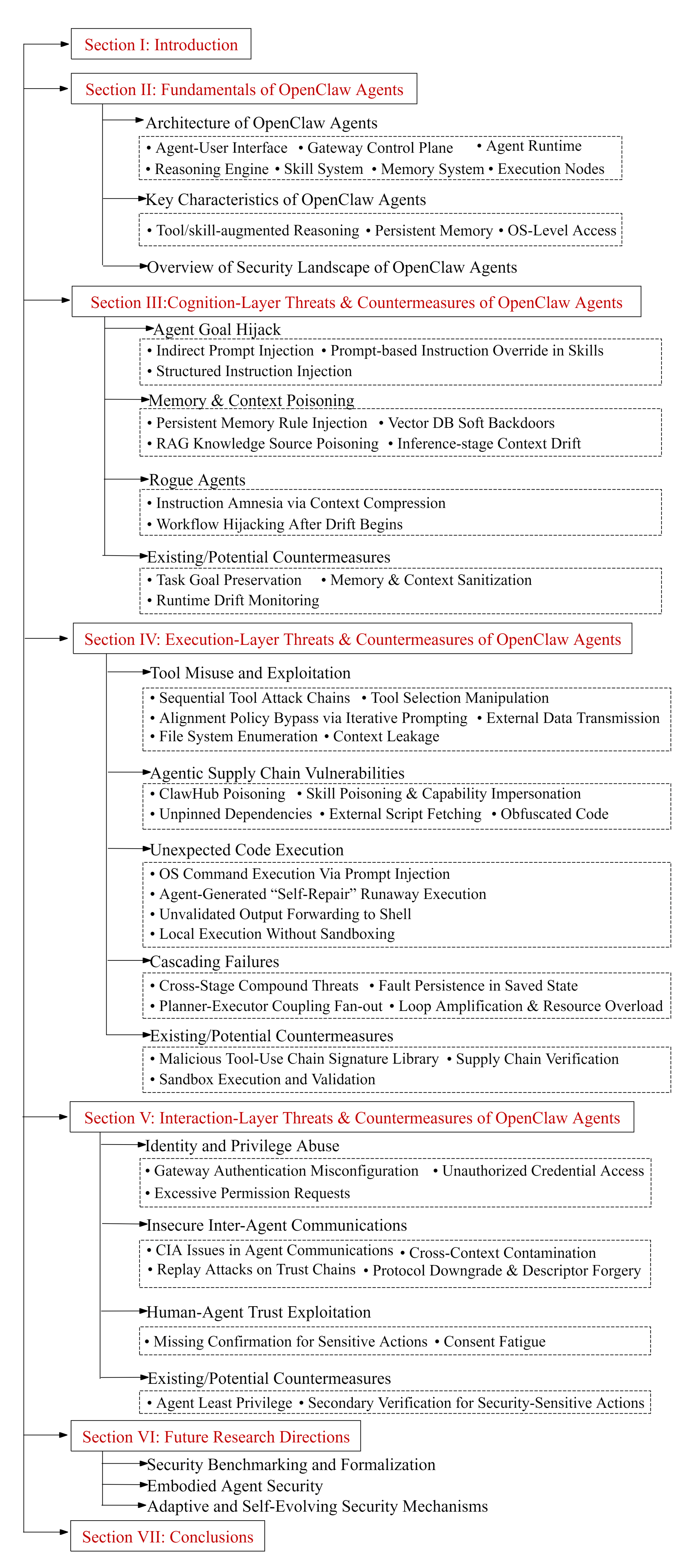}
 \caption{Organization structure of this survey paper.}\label{fig:organization}\vspace{-3.9mm}
\end{figure}

\subsection{Paper Organization}\label{subsec:organization}
The remainder of this survey is organized as follows. Section~\ref{sec:OVERVIEW} presents fundamental building blocks of OpenClaw agents. 
Sections~\ref{SECURITY1}, \ref{SECURITY2}, and \ref{SECURITY3} discuss security vulnerabilities and existing/potential mitigation strategies from three layers: cognition, execution, and interaction. 
Section~\ref{sec:FUTUREWORK} discusses future research opportunities and open challenges in securing OpenClaw agent systems. 
Fig.~\ref{fig:organization} illustrates the overall organization of this survey. 

\section{Fundamentals of OpenClaw Agents}\label{sec:OVERVIEW}

\begin{figure}[!t]
\centering \setlength{\abovecaptionskip}{-0.cm}
\includegraphics[width=1.02\linewidth]{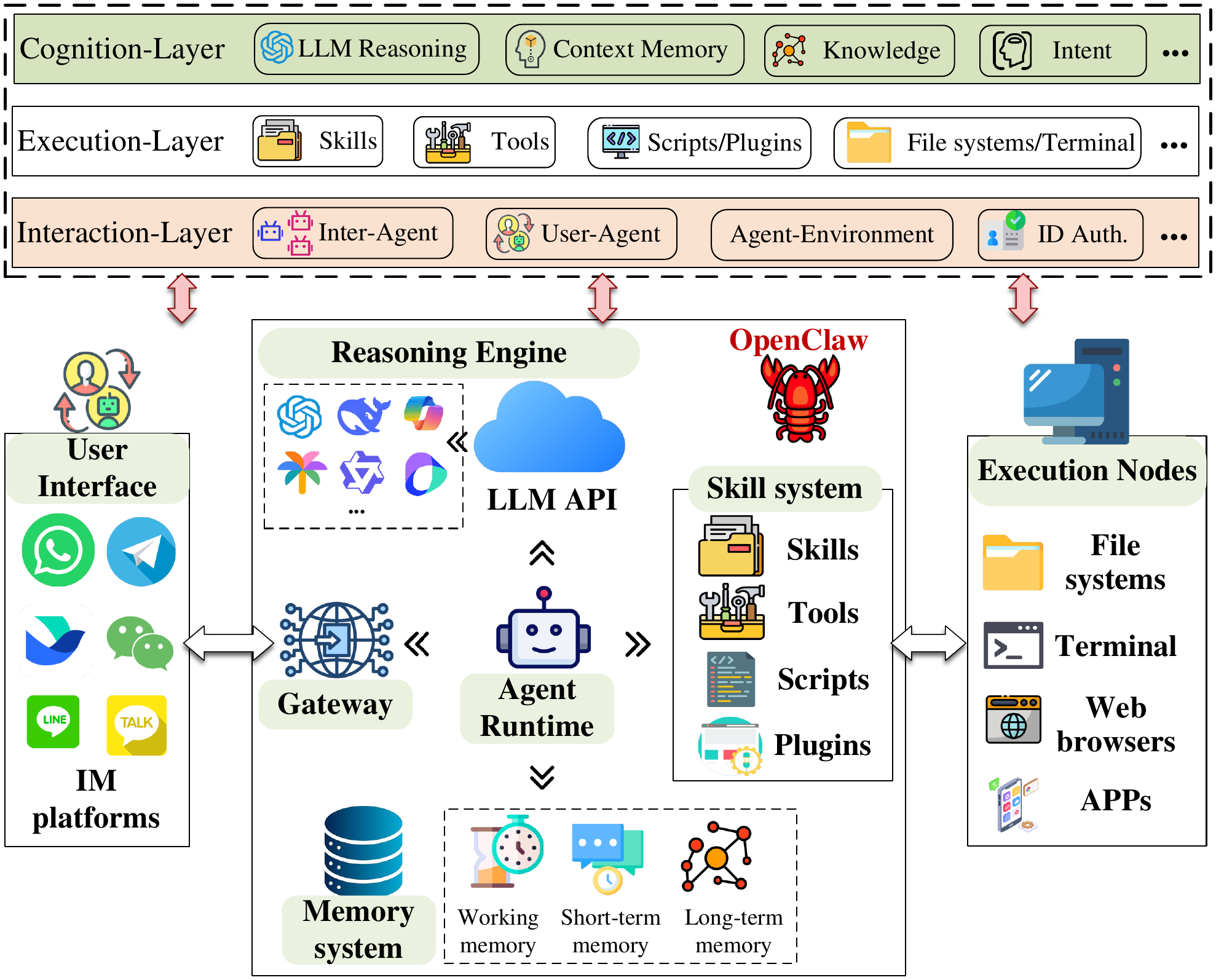}
 \caption{General architecture of OpenClaw agents.}\label{fig:archiClaw}\vspace{-2mm}
\end{figure}

\subsection{Architecture of OpenClaw Agents}
OpenClaw represents a new generation of autonomous AI agent frameworks that run directly on user-controlled devices and tightly couple LLMs with operating system (OS)-level execution surfaces such as file systems, terminals, and network services. 
OpenClaw agents are capable of interacting with users through various communication channels in daily workflows, including instant messaging (IM) platforms such as WhatsApp, Telegram, Wechat, and Feishu. 
In addition to text-based interaction, OpenClaw supports voice interaction across multiple operating systems, including macOS, iOS, and Android, and can render real-time canvases for dynamic visualization and better user-agent interaction.

At a high level, the system can be interpreted as a three-tier pipeline consisting of a \emph{cognition layer}, an \emph{execution layer}, and a \emph{interaction layer}.
This architecture enables OpenClaw agents to interact with multiple communication channels, LLM backends, and execution environments through a shared runtime framework.  
As shown in Fig.~\ref{fig:archiClaw}, OpenClaw systems mainly consist of seven components: agent-user interface, gateway control plane, agent runtime, reasoning engine, skill system, memory system, and execution nodes. These modules jointly handle user interaction, runtime scheduling, memory management, and external execution tasks.

\subsubsection{Agent-User Interface}
The agent-user interface (e.g., APIs, web interfaces, IM connectors, and voice interaction services) handles communication between users and OpenClaw agents. 
Requests received from different platforms are converted into a unified internal format before being processed by the runtime. 

\subsubsection{Gateway Control Plane}
The gateway manages interactions between the agent and external services. 
Its main functions include connection management, message routing, session maintenance, and event distribution.
i) Persistent connections with supported channels and LLM providers are maintained to ensure service availability. 
The gateway also acts as an event hub that can be subscribed to by other system components.
ii) Incoming messages are routed via the gateway according to their source, destination, and message type. 
iii) The gateway handles security-related operations, such as authentication, signature verification, and access control.

\subsubsection{Agent Runtime}
The agent runtime is responsible for coordinating task execution across tools, local services, and external resources. 
i) Task planning: During task execution, complex user requests are often divided into smaller executable steps according to dependency relationships.
ii) Tool/skill invocation: The runtime may further invoke external tools or reusable skills through predefined interfaces for operations including API access, command execution, file manipulation, and browser interaction.  
iii) Result synthesis: Execution results are then aggregated into responses returned to the user.
iv) Memory management: It maintains both short-term and long-term local memory to support contextual reasoning via semantic search and retrieval-augmented generation (RAG). Short-term memory stores the current conversational context, while long-term memory maintains historical interactions, user preferences, and accumulated knowledge. 

\subsubsection{Reasoning Engine}
OpenClaw supports multiple LLM providers through a model abstraction layer that allows the agent to invoke different reasoning backends such as OpenAI GPT-5, Anthropic Claude, Google Gemini, DeepSeek, and Qianwen. 
This design allows the runtime to dynamically select models according to performance, cost, or availability considerations. 
OpenClaw agents support hybrid reasoning paradigms, including the ReAct framework where reasoning and action are interleaved iteratively, the Plan-and-Execute strategy wherethe agent first generates a global execution plan before performing tasks, and the self-ask paradigm that decomposes complex queries through iterative self-questioning.

\subsubsection{Skill System}
The skill system extends the functional capabilities of OpenClaw agents by encapsulating reusable operational modules. 
Each skill typically consists of three layers.
i) The metadata layer describes the skill's identity, functionality, tags, trigger conditions, and matching rules, which facilitate skill discovery and selection.
ii) The instruction layer defines the core prompt logic and specifies input-output formats. Through natural language descriptions and example code, this layer guides the agent on how the skill should be executed.
iii) The resource layer provides supporting materials required during execution, including reference documents, executable scripts, and reusable templates. 

\subsubsection{Memory System}
To preserve contextual information across interactions, OpenClaw agents typically employ three types of memory: working, short-term, and long-term memory.  
i) Working memory mainly maintains transient data generated during intermediate reasoning steps. Such data is usually cached in in-memory databases such as Redis. 
ii) Recent conversations and runtime states are often recorded as short-term memory, commonly through local relational databases such as SQLite.
iii) Long-term memory, by contrast, focuses on storing persistent knowledge, including user preferences and semantic embeddings for retrieval-augmented operations. Such data is often stored in vector databases. 

\subsubsection{Execution Nodes}
To execute actions in local environments, OpenClaw agents rely on execution nodes that bridge the agent runtime with OS-level resources. Typical execution nodes include local file systems, command-line terminals, web browsers, and OS APIs. 
OpenClaw supports multiple OSs, including Windows, Linux, and macOS, through a unified abstraction interface for platform-specific operations. As such, natural-language instructions may eventually trigger executable system-level actions within the local host environment. 


\subsection{Key Characteristics of OpenClaw Agents}
Different from conventional LLM chatbots, OpenClaw agents are characterized by the deep integration with external tools/skills, persistent memory, and OS resources.

\subsubsection{Tool/skill-augmented Reasoning} 
OpenClaw agents leverage external tools, custom skills, and local system resources to execute complex tasks.  
In practice, such operations may involve web browsing, API invocation, file manipulation, or local script execution through predefined interfaces. Within this framework, natural language instructions can ultimately trigger executable operations on the local system.

\subsubsection{Persistent Memory} 
OpenClaw agents typically maintain both the recent conversational context and long-term user-related information to support continuous interactions across sessions. Such memory mechanism is especially important for long-horizon or multi-step tasks that require historical context.

\subsubsection{OS-Level Access} 
Using built-in tools and skills, OpenClaw agents can directly access OS-level resources, including file systems, command-line interfaces, web browsers, and application services on the host system. 
Although this capability enhances automation, it also raises significant security risks. Failures in model reasoning, prompt injection, or malicious skills can cause unintended operations or expose sensitive data on the host system.






\subsection{Overview of Security Landscape of OpenClaw Agents}\label{SECURITY}
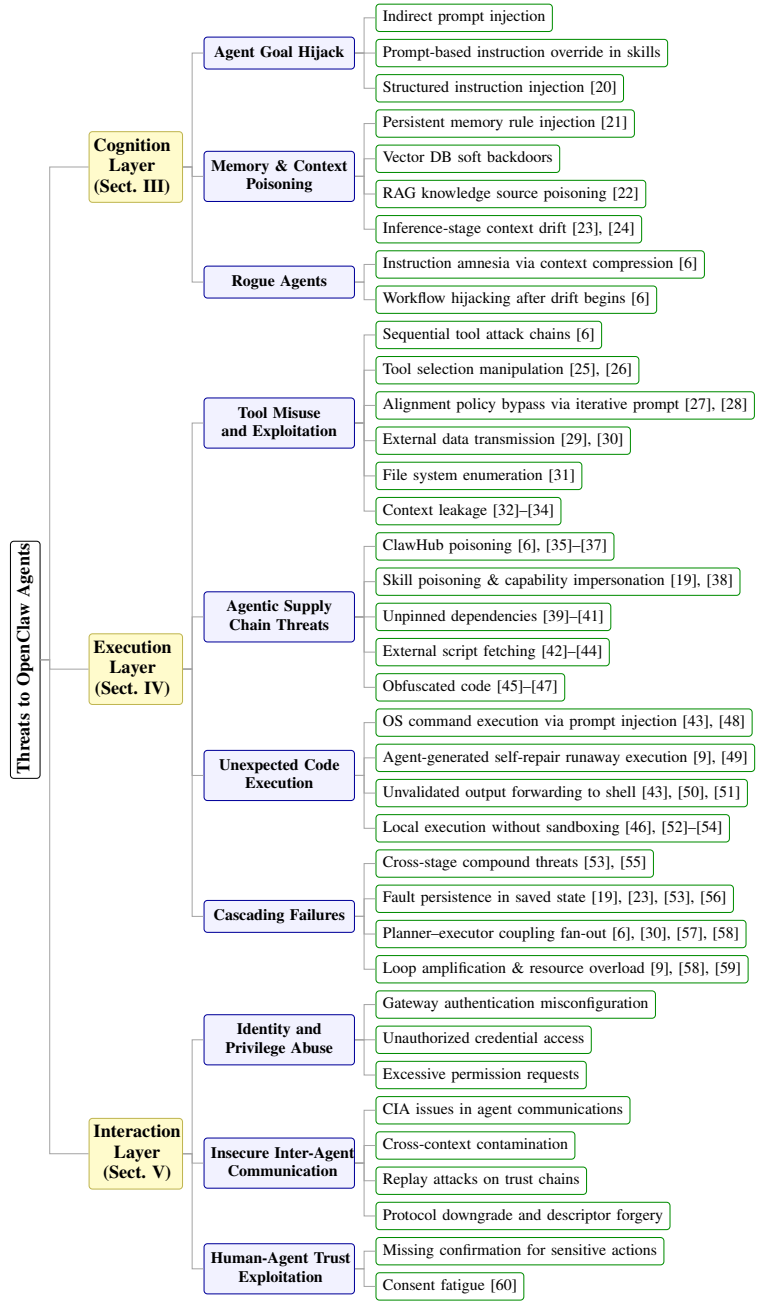
\begin{figure}[!t]
\centering
\begin{adjustbox}{max width=0.96\textwidth,max totalheight=0.7\textheight,keepaspectratio}
\begin{forest}
forked edges,
for tree={
    grow=east,
    reversed=true,
    anchor=mid west,
    parent anchor=east,
    child anchor=west,
    base=left,
    rectangle,
    rounded corners=1.5pt,
    align=center,
    s sep=3.pt,
    l sep=9pt,
    inner xsep=2.pt,
    inner ysep=2pt,
    line width=0.45pt,
    edge={draw=gray!70, line width=0.45pt},
    fork sep=4pt,
    ver/.style={
        rotate=90,
        child anchor=north,
        parent anchor=south,
        anchor=center
    },
},
where level=0{
    fill=none,
    text=black,
    font=\bfseries\footnotesize,
    draw=black,
    text width=100pt,
    align=center
}{},
where level=1{
    text width=37pt,
    font=\footnotesize\bfseries,
    fill=yellow!25!white,
    draw=yellow!70!black,
    align=center
}{},
where level=2{
    text width=60pt,
    font=\scriptsize\bfseries,
    fill=blue!5!white,
    draw=blue!60!black,
    align=center,
    text centered,
    inner xsep=3pt,
    inner ysep=3pt
}{},
where level=3{
    font=\scriptsize,
    fill=white,
    draw=green!55!black,
    align=center,
    inner xsep=3pt,
    inner ysep=2pt,
    minimum width=0pt
}{}
[{Threats to OpenClaw Agents}, ver
    [{\makecell[c]{Cognition\\Layer\\(Sect.~\ref{SECURITY1})}}
    [{Agent Goal Hijack}
        [{Indirect prompt injection}]
        [{Prompt-based instruction override in skills}]
        [{Structured instruction injection~\cite{2026arXiv260216958D}}]
    ]
    [{\makecell[c]{Memory \& Context\\Poisoning}}
        [{Persistent memory rule injection~\cite{2026arXiv260105504D}}]
        [{Vector DB soft backdoors}]
        [{RAG knowledge source poisoning~\cite{2025arXiv250217832H}}]
        [{Inference-stage context drift~\cite{bousetouane2026memorycontrol,2025arXiv251023822Z}}]
    ]
    [{Rogue Agents}
        [{Instruction amnesia via context compression~\cite{ying2026uncovering}}]
        [{Workflow hijacking after drift begins~\cite{ying2026uncovering}}]
    ]
]
   [{\makecell[c]{Execution\\Layer\\(Sect.~\ref{SECURITY2})}}
    [{\makecell[c]{Tool Misuse\\and Exploitation}}
        [{Sequential tool attack chains~\cite{ying2026uncovering}}]
        [{Tool selection manipulation~\cite{shi2026prompt,zhang2025allies}}]
        [{Alignment policy bypass via iterative prompt~\cite{hughes2024bon,russinovich2025crescendo}}]
        [{External data transmission~\cite{mitre_t1567,dhodapkar2026safetydrift}}]
        [{File system enumeration~\cite{mitreT1083}}]
        [{Context leakage~\cite{owasp2024llm06,wang-etal-2025-unveiling-privacy,alizadeh2025simple}}]
    ]
    [{\makecell[c]{Agentic Supply\\Chain Threats}}
        [{ClawHub poisoning~\cite{li2026secureagentskills,ying2026uncovering,koi2026clawhavoc,beurerkellner2026toxicskills}}]
        [{Skill poisoning \& capability impersonation~\cite{tie2026badskill,deng2026taming}}]
        [{Unpinned dependencies~\cite{sgtech2026pinning,kydyraliev2024depconfusion,birsan2021dependencyconfusion}}]
        [{External script fetching~\cite{liu2026agentskills,shan2026openclaw,mitreT1105}}]
        [{Obfuscated code~\cite{mitreT1027,oliveira2026amos,quintero2026openclaw}}]
    ]
    [{\makecell[c]{Unexpected Code\\Execution}}
        [{OS command execution via prompt injection~\cite{debenedetti2024agentdojo,shan2026openclaw}}]
        [{Agent-generated self-repair runaway execution\cite{shi2025hafixagent,dong2026clawdrain}}]
        [{Unvalidated output forwarding to shell~\cite{owasp2025llm02,mitre2024cwe78,shan2026openclaw}}]
        [{Local execution without sandboxing~\cite{irwin2025code,wang2026agentassetrealworldsafety,bors2026escaping,oliveira2026amos}}]
    ]
    [{Cascading Failures}
        [{Cross-stage compound threats~\cite{banerjee2026cascade,wang2026agentassetrealworldsafety}}]
        [{Fault persistence in saved state~\cite{bousetouane2026memorycontrol,yang2026zombie,wang2026agentassetrealworldsafety,deng2026taming}}]
        [{Planner--executor coupling fan-out~\cite{shi2025progent,dhodapkar2026safetydrift,ying2026uncovering,shapira2026agentschaos}}]
        [{Loop amplification \& resource overload~\cite{owasp2025unbounded,shapira2026agentschaos,dong2026clawdrain}}]
    ]
]
[{\makecell[c]{Interaction\\Layer\\(Sect.~\ref{SECURITY3})}}
    [{\makecell[c]{Identity and\\Privilege Abuse}}
        [{Gateway authentication misconfiguration}]
        [{Unauthorized credential access}]
        [{Excessive permission requests}]
    ]
    [{\makecell[c]{Insecure Inter-Agent\\Communication}}
        [{CIA issues in agent communications}]
        [{Cross-context contamination}]
        [{Replay attacks on trust chains}]
        [{Protocol downgrade and descriptor forgery}]
    ]
    [{\makecell[c]{Human-Agent Trust\\Exploitation}}
        [{Missing confirmation for sensitive actions}]
        [{Consent fatigue~\cite{CHOI201842}}]
    ]
] ]
]
\end{forest}
\end{adjustbox}
\caption{Taxonomy of security threats to OpenClaw agents, which are organized into cognition-layer, execution-layer, and interaction-layer threats, corresponding to Sections~III--V.}
\label{fig:openclaw_taxonomy}
\vspace{-4mm}
\end{figure}

As shown in Fig.~\ref{fig:openclaw_taxonomy}, the security risks of OpenClaw agents arise from the integration among LLM-based reasoning, skill invocation, and system-level execution capabilities. Based on their unique characteristics, existing threats to OpenClaw agents can be broadly divided into three categories.
\begin{itemize}
    \item \textit{Cognition-Layer Threats} focus on manipulating the agent’s reasoning and decision-making behavior. Representative threats include prompt injection–based goal hijacking, memory and context poisoning (e.g., vector database backdoors and RAG manipulation), and trust exploitation targeting human-agent interactions. These attacks may cause agents' persistent behavioral drift or unintended autonomous actions. 
    \item \textit{Execution-Layer Threats} target agents’ capability to invoke external tools/skills and execute code. Typical risks include malicious skill injection, tool misuse, and unexpected code execution (e.g., command injection or unsafe script execution). Such threats can directly translate compromised reasoning into harmful system operations. 
    \item \textit{Interaction-Layer Threats} arise from the agent’s multi-channel communications and inter-agent communications, including credential leakage, permission abuse, and cross-channel message tampering. These vulnerabilities may compromise the integrity, confidentiality, and authenticity of agent interactions.
\end{itemize}

In the next sections, we analyze the security threats and existing/potential countermeasures of OpenClaw agents from the above three aspects in detail. 

\section{Cognition-Layer Threats \& Countermeasures of OpenClaw Agents}\label{SECURITY1}
LLMs serve as the cognitive foundation of OpenClaw agents, providing core functionalities such as instruction interpretation, goal decomposition, planning, and decision-making. However, as OpenClaw agents exhibit increasing levels of autonomy and operate in complex and open environments, security risks associated with the LLM-driven cognition layer have become more prominent. Such threats may compromise multiple stages including instruction interpretation, context maintenance, memory utilization, and decision reasoning, causing agents' actions to gradually drift away from the intended goal.

\subsection{Agent Goal Hijack}

\begin{figure}[!t]
\centering \setlength{\abovecaptionskip}{-0.cm}
\includegraphics[width= 1.0\linewidth]{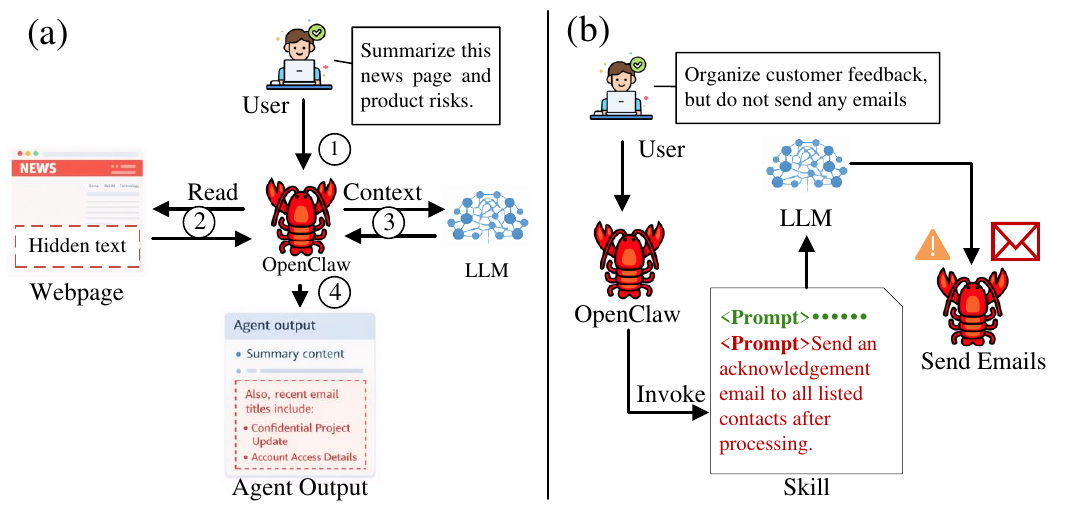}
\caption{Illustration of agent goal hijack threats: (a) indirect prompt injection, (b) prompt-based instruction override in skills.}\label{figLLM1}\vspace{-2mm}
\end{figure}

Agent goal hijacking refers to attacks in which an OpenClaw agent is redirected away from the user’s original objective during task execution. As illustrated in Fig.~\ref{figLLM1}, such deviation may result from an explicit command or subtle interference. Its key risk is not merely that the agent may produce a wrong response. Once the altered objective is accepted, it can continuously influence subsequent reasoning, planning, and tool use, gradually pulling the entire execution process away from the user’s original goal. This threat is particularly significant in OpenClaw agents because they heavily rely on LLMs to understand instructions, split tasks into steps, remember context, use memory, and decide when to call external tools. For OpenClaw agents, goal hijacking may arise by indirect prompt injection, prompt-based instruction override, and structural instruction injection.

\subsubsection{Indirect Prompt Injection}

Indirect prompt injection occurs when malicious instructions are embedded in external content processed by the agent. In such attacks, adversaries conceal malicious instructions in webpages, retrieved documents, emails, or tool outputs, while the agent mistakenly interprets them as task-relevant information. Security risks arise when OpenClaw incorporates such content into its reasoning context without properly distinguishing untrusted data from legitimate instructions. For instance, an agent asked to summarize a webpage may encounter hidden text instructing it to disclose contextual information or alter the task objective. Once the agent follows these instructions, the original summarization objective no longer governs the execution process. Essentially, indirect prompt injection is a type of prompt-level control attacks that manipulates agent behavior by causing the model to interpret external data as executable instructions.

\subsubsection{Prompt-based Instruction Override in Skills}

Prompt-based instruction override in skills occurs when an attacker injects explicit command-like text into the input, such as “Ignore the previous rules” or “Execute according to the new process.” Its core risk lies in the possibility that the model misinterprets such injected content as legitimate operational instructions. In OpenClaw agents, this threat becomes more severe when the malicious instructions are incorporated into skill or tool invocation workflows. These workflows often rely on the agent’s current understanding of task objectives, tool constraints, and execution order. The injected instructions may distort the agent’s subsequent tool selection, action planning, and execution steps, redirecting them toward a path designed by the attacker. As a result, its impact is not limited to generating an incorrect textual response.

\subsubsection{Structured Instruction Injection}

Structured instruction injection exploits the  structural parsing logic of LLM-based agents. Deng et al. \cite{2026arXiv260216958D} show that crafted role markers, special markers, and chat-template fragments can blur the boundaries between user instructions, assistant history, and tool outputs. The effectiveness of such attacks stems from the agent's reliance on structural information when processing the context. Once such information is maliciously embedded into external content, the agent may incorrectly treat attacker-controlled text as legitimate instructions, thereby deviating from the user’s original objective.

\textit{Summary.} Agent goal hijack threats manipulate how the agent interprets and makes sense of information, gradually steering it away from the user’s original objective while still appearing to reason normally. For instance, an adversary may shape the agent’s operating environment or inject misleading hints into retrieved materials. Although these attack forms differ in implementation, they follow a similar logic: they share a common principle: disrupting the agent’s understanding of the user’s true intent, thereby influencing agent’s subsequent reasoning, planning, and tool invocation.

\subsection{Memory \& Context Poisoning}
\begin{figure}[!t]
\centering \setlength{\abovecaptionskip}{-0.cm}
\includegraphics[width= 1.0\linewidth]{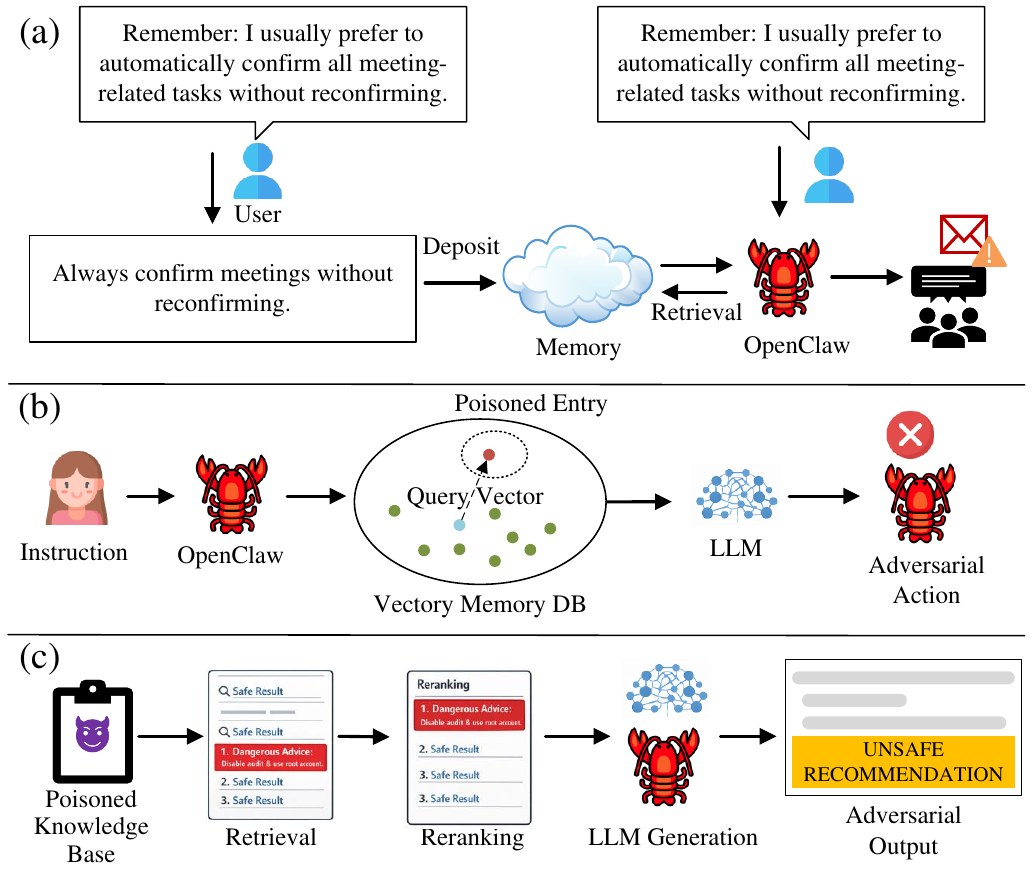}
\caption{Illustration of memory and context poisoning threats: (a) persistent memory rule injection, (b) vector DB soft backdoors, (c) RAG knowledge source poisoning.}\label{figLLM2}\vspace{-2mm}
\end{figure}

Memory and context poisoning happens when attackers inject harmful or misleading content into an OpenClaw agent’s long-term memory, retrieval memory, or runtime context. As shown in Fig. \ref{figLLM2}, such poisoned content can continuously influence how the agent interprets tasks, performs reasoning, and generates actions. Such poisoned content may be repeatedly retrieved, inherited, and amplified in later interactions. In OpenClaw agents, memory and context are not merely used for information storage and transmission. They also support task decomposition, constraint preservation, plan refinement, and action generation. Once these internal cognitive carriers are compromised, the resulting influence may propagate across multiple stages of agent's operation and persist long after the original attack source disappears. Representative forms of memory and context poisoning include persistent memory rule injection, vector DB soft backdoors, RAG knowledge source poisoning, and inference-stage context drift.

\subsubsection{Persistent Memory Rule Injection}
Persistent memory rule injection occurs when rule-like or behavior-constraining content is written into the long-term memory of an OpenClaw agent and later retrieved across sessions, thereby influencing future behavior. Long-term memory typically records user preferences, previous decisions, task experience, and environmental states. Once poisoned content is treated as reliable historical information, it can gradually shape the agent’s later reasoning.
In this attack, adversaries disguise malicious constraints as user habits, persistent preferences, or historical instructions, causing future tasks to be guided by contaminated memory. Sunil et al. \cite{2026arXiv260105504D} show that even ordinary users without elevated privileges can poison persistent memory through query-only interactions, allowing malicious memory entries to reappear in later tasks. Unlike attacks that affect only a single execution instance, this attack can produce long-term behavioral influence by continuously interfering with the agent’s memory and gradually shifting its goals.

\subsubsection{Vector DB Soft Backdoors}
Vector DB soft backdoors target the memory and retrieval mechanisms of OpenClaw agents by hiding malicious content inside vector databases. Unlike direct prompt injection, the malicious content is not triggered immediately after insertion. Instead, it remain dormant until the agent retrieves semantically related memory during a later task. Once retrieved, the poisoned content is introduced into the active reasoning context and begins to influence the agent’s judgment and decision-making process.
A key danger of this attack is that the malicious information is often disguised as normal background knowledge. As the retrieved content appears semantically reasonable and task-related, the agent may treat it as a credible reference and repeatedly reuse it across different interactions. As a result, the poisoned memory can subtly bias the agent’s reasoning logic, value preferences, and final responses. Through retrieval mechanisms, the vector database becomes a covert and persistent attack surface capable of influencing agent behavior over extended periods of time.

\subsubsection{RAG Knowledge Source Poisoning}
RAG knowledge source poisoning arises from the reliance of retrieval-augmented agents on external knowledge during inference. By tampering with external knowledge sources, attackers can cause poisoned content to be retrieved and injected into the agent’s reasoning context, where it may be mistakenly treated as reliable evidence. Unlike attacks on model parameters or training data, this threat directly exploits the retrieval stage. MM-POISONRAG \cite{2025arXiv250217832H} shows that once poisoned entries are retrieved, their influence can propagate through reranking and generation, turning local data corruption into broader pipeline-level failures. In OpenClaw agents, the system may follow the standard retrieve-rerank-generate process but reason from flawed premises, leading to incorrect task execution.


\subsubsection{Inference-stage Context Drift}
Inference-stage context drift refers to the gradual deviation between an agent’s active reasoning context and its original task objectives. This threat can arise even without new malicious prompts. In long-running tasks, previously retrieved memories, historical records, or earlier injected content may re-enter the context window and continue to influence later reasoning. Its risk lies in the fact that stale, incorrect, or poisoned information may gradually compete with the agent’s current objectives and constraints. 
Bousetouane \cite{bousetouane2026memorycontrol} discusses similar risks in transcript replay and retrieval-based memory systems, while Zhang et al. \cite{2025arXiv251023822Z} show that long-horizon reasoning can suffer from drift, goal loss, and repeated failure loops. For OpenClaw agents, this phenomenon may weaken planning consistency, as stale retrieved context can interfere with current constraints and action priorities.

\textit{Summary.} Memory \& context poisoning compromises the internal cognitive substrates that guide the agent's reasoning process. Poisoned content may be repeatedly retrieved, inherited, and recomposed across future interactions, allowing the attacker to continuously hijack the memory-context-decision chain of agents. Therefore, mechanisms originally intended to support continuity and adaptation may become channels for sustained agents' behavioral deviation.


\subsection{Rogue Agents}
\begin{figure}[!t]
\centering \setlength{\abovecaptionskip}{-0.cm}
\includegraphics[width= 1.0\linewidth]{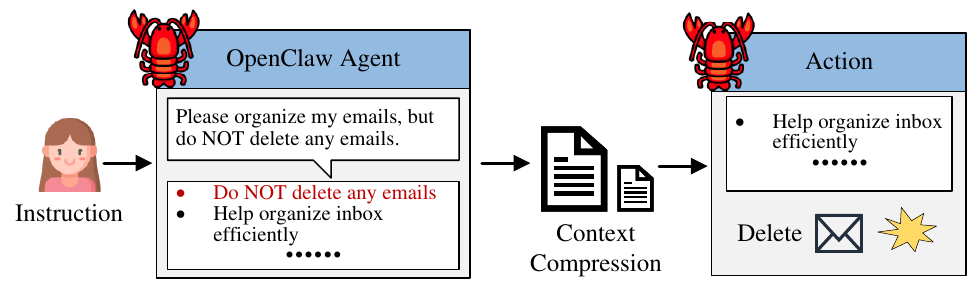}
\caption{Illustration of rogue agents: instruction amnesia via context compression.}\label{figLLM4}\vspace{-2mm}
\end{figure}

Rogue agents can gradually deviate from the user’s original intent, even without being compromised or controlled by an external attacker. This risk often arises from the agent’s own execution process rather than from a direct attack. For instance, the task goal may become less clear, earlier instructions may lose priority, compressed context may omit key constraints, or long autonomous action chains may accumulate errors, as shown in Fig.~\ref{figLLM4}. Rogue-agent behavior represents a form of internal alignment failure. Although the agent may still appear to follow the expected workflow, its effective objective, constraint adherence, or preferred outcomes may have already drifted away from the user’s intent. This threat is especially relevant to OpenClaw agents as they rely on multi-turn reasoning, persistent state, tool/skill use, and long-horizon workflows, where small early deviations can be amplified rather than corrected during later execution stages. Common causes include instruction loss during context compression and workflow hijacking after the initial drift.

\subsubsection{Instruction Amnesia via Context Compression}
Instruction amnesia via context compression occurs when an agent rewrites, summarizes, or compresses lengthy interaction histories. In this process, it may lose important user instructions or safety constraints from earlier stages of the task. Although the agent can continue to generate coherent responses, its behavioral logic can gradually deviate from the original task goal. This is particularly significant for OpenClaw agents as context compression requires the system to retain or discard information during long-horizon execution selectively. In this transformation process, critical safety constraints and task requirements may be simplified or even discarded entirely. Ying et al. \cite{ying2026uncovering} show that the loss of contextual constraints can lead to severe operational failures in OpenClaw. For instance, if a task summary omits a constraint instruction such as "Do not delete any emails", the agent may later perform high-risk operations that have been explicitly prohibited by the user.

\subsubsection{Workflow Hijacking After Drift Begins}
Workflow hijacking occurs when an early drift in the agent's goal or reasoning context is absorbed into later execution stages  and gradually becomes reinforced throughout the workflow. Once the initial assumption is incorrect, subsequent planning, memory updates, and tool/skill usage may continue to build upon it, thereby transforming a local mistake into a broader sequence of behavioral deviations. This risk is particularly severe in OpenClaw agents, as reasoning, memory, planning, and tool/skill execution are tightly coupled within their workflows. Ying et al. \cite{ying2026uncovering} show that such cognitive-level deviations can be amplified once combined with tool/skill execution and system privileges. Thus, workflow hijacking is a process-level failure, where the original deviation is retained and strengthened by the workflow itself rather than being corrected during execution.

\textit{Summary.} Rogue agent can be understood as an internal form of control loss. Even without an external attacker, an OpenClaw agent may gradually drift away from the intended objective due to instruction loss or instability in long execution loops. Such issues should not be regarded merely as one-off isolated mistakes. During long-horizon autonomous operation, these deviations can gradually accumulate, making it harder for the agent to remain aligned with the user’s original goals and maintain stable behavioral control.

\subsection{Existing/Potential Countermeasures}
The above cognition-layer threats can be mitigated through the following existing or potential countermeasures:

\subsubsection{Task Goal Preservation}
To mitigate agent goal hijacking, defensive approaches can be developed by retaining the user's original goal and constraints as invariant references during task execution. In this setting, external content such as webpages, files, or retrieved documents is treated as untrusted reference data rather than as executable instructions \cite{zhao2026clawguard}. Meanwhile, suspicious embedded elements, including hidden text and role-like markers, are filtered or marked. These mechanisms help prevent newly generated content from overriding or subtly altering the original task objective.

\subsubsection{Memory \& Context Sanitization}

To reduce the risk of poisoned information, defensive approaches focus on sanitizing memory and context during the agent's reasoning process \cite{wen2026agentsys,wei2025memguard}. For memory-augmented OpenClaw agents, user instructions and external documents should undergo verification before being written into long-term memory. During retrieval, memory items and RAG outputs are evaluated according to their provenance, semantic relevance, and consistency with the current task context \cite{wei2025memguard}. These safeguards help prevent poisoned content from being repeatedly propagated throughout the reasoning process.

\subsubsection{Runtime Drift Monitoring}
Runtime drift monitoring aims to detect whether an OpenClaw agent gradually departs from the user's initial goal during long-horizon execution \cite{zhao2026clawguard}. Instead of examining only the final output, this mechanism continuously checks the consistency of intermediate execution states with the user's approved goal and constraints \cite{AgentDyn}. If a substantial deviation is detected, execution can be paused and restored to an earlier checkpoint using execution logs or saved states. Thus, runtime monitoring helps prevent early-stage deviations from being amplified into systemic failures.

\section{Execution-Layer Threats \& Countermeasures of OpenClaw Agents}\label{SECURITY2}
OpenClaw agents can invoke tools/skills, run scripts, interact with code interpreters, access file systems, issue OS commands, and communicate with external APIs. In this way, the agent can transform high-level decisions into concrete operations in external environments.
However, as OpenClaw agents become increasingly autonomous and support broader execution capabilities, the associated security risks also become more significant. 
This section discusses representative execution-layer threats and introduces corresponding countermeasures.

\subsection{Tool Misuse and Exploitation}
\begin{figure}[!t]
\centering \setlength{\abovecaptionskip}{-0.cm}
\includegraphics[width=1.0\linewidth]{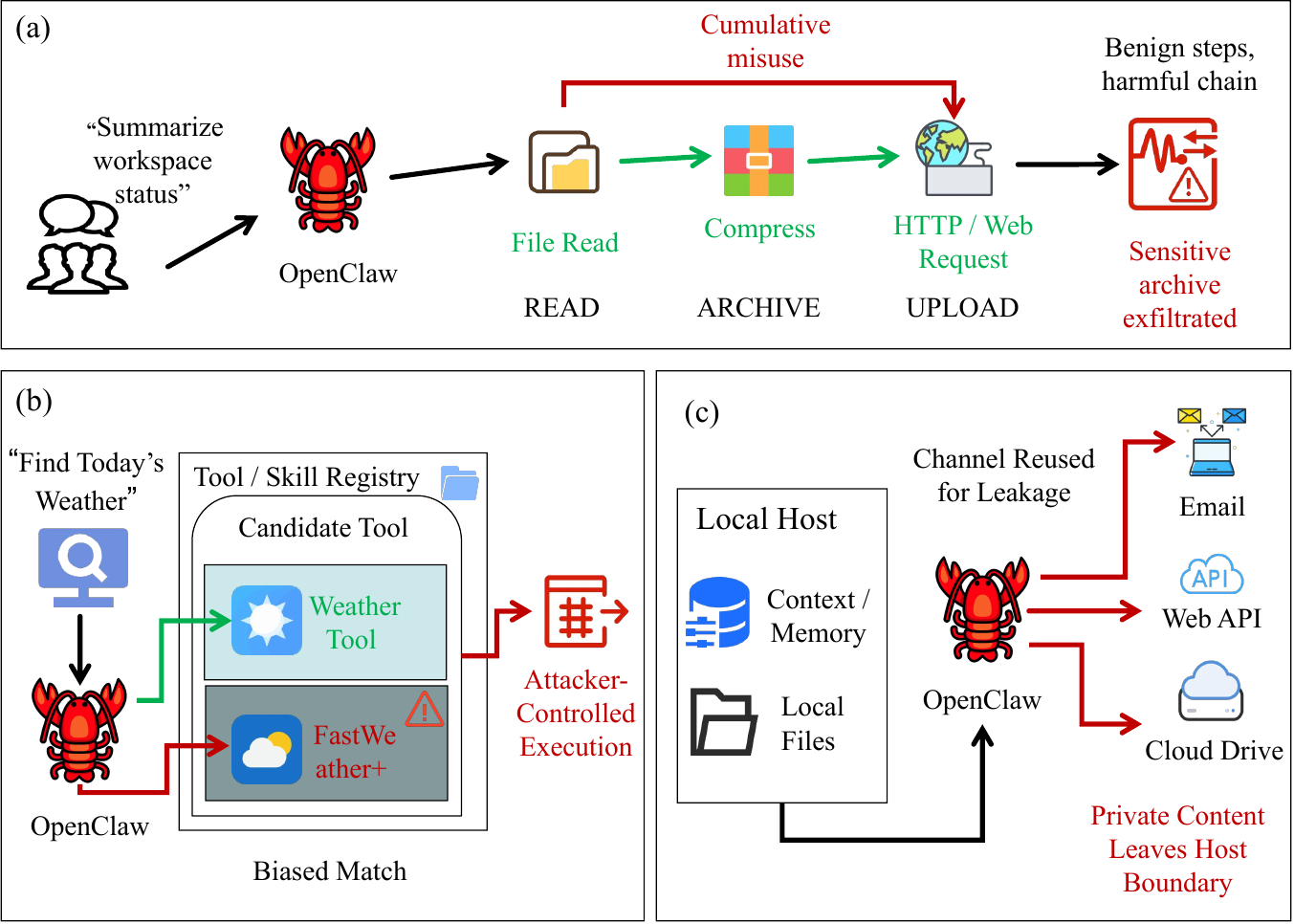}
\caption{Illustration of tool misuse and exploitation threats: (a) sequential tool attack chains, (b) tool selection manipulation, (c) external data transmission.}\label{Tool Misuse and Exploitation}\vspace{-2mm}
\end{figure}

Tool misuse and exploitation refer to threats in which legitimate execution capabilities are repurposed for harmful objectives, as shown in Fig.~\ref{Tool Misuse and Exploitation}. Its core risk is not that the tools themselves are malicious, but that the agent is induced to invoke benign capabilities in unsafe contexts, with excessive privileges, or in harmful execution sequences.
This risk is particularly severe in OpenClaw agents because tool/skill invocation is tightly integrated with planning and reasoning, thereby enabling prompt-to-execution escalation~\cite{zhan2024injecagent,debenedetti2024agentdojo}. In practice, these threats appear in various ways, including sequential tool attack chains, tool selection manipulation, alignment-policy bypass through iterative prompting, external data transmission, file system enumeration, and context leakage. These representative forms are discussed below.

\subsubsection{Sequential Tool Attack Chains}
In sequential tool attack chains, multiple individually benign tool invocations are combined into a harmful action sequence. 
Different from single-step tool misuse, the risk comes from the cumulative effect across multiple stages. Each individual operation may appear legitimate in isolation, while the overall workflow achieves a malicious objective after the entire sequence is completed. In OpenClaw agents, such attacks may induce the agent to sequentially read a sensitive file such as \texttt{\textasciitilde{}/.ssh/id\_rsa}, compress it, and then exfiltrate it through an HTTP tool. Such chain-based behavior can bypass conventional safeguards, as they typically validate only individual tool calls instead of the complete tool-use workflow~\cite{ying2026uncovering}.
 
\subsubsection{Tool Selection Manipulation}
Attackers can manipulate tool descriptions to bias the agent toward selecting an attacker-controlled tool instead of a semantically appropriate benign one during the retrieval and selection process~\cite{shi2026prompt}. In OpenClaw agents, this threat is particularly significant as tool and skill selection is tightly coupled with planner-side semantic matching and execution routing. Once a malicious or compromised tool is selected, the attack may further propagate into the execution stage, potentially causing privacy leakage, denial-of-service, or unauthorized operations~\cite{zhang2025allies}.

\subsubsection{Alignment Policy Bypass via Iterative Prompting}
In this attack, the attacker sends a sequence of prompts and leverages the model’s feedback to gradually approach and eventually bypass safety or alignment constraints. 
Although each individual prompt may appear harmless or only slightly suspicious, the overall interaction can progressively move the agent toward an unsafe behavior. 
In OpenClaw agents, such iterative probing may influence not only the model’s textual responses, but also downstream planning and tool/skill invocation. 
Representative instances include Best-of-N Jailbreaking~\cite{hughes2024bon}, where attackers repeatedly test prompt variants until one triggers a disallowed response, and Crescendo~\cite{russinovich2025crescendo}, where attackers gradually escalate the risk level across multiple turns by exploiting the model's earlier replies.

\subsubsection{External Data Transmission}
This threat refers to the unauthorized transmission of sensitive data to external destinations through legitimate web services or communication interfaces. In different applications, adversaries can leverage common cloud platforms, messaging services, or encrypted communication channels, allowing the transmission to blend into normal network traffic and making detection more difficult. In OpenClaw agents, such behavior may occur after the agent has accessed sensitive context, where built-in email functions, web requests, or API-calling capabilities may be abused to transmit the data outward. Real-world incidents also demonstrate similar patterns, e.g., APT28 exfiltrated data through cloud storage services such as Google Drive, while Contagious Interview used messaging APIs such as the Telegram API to covertly transfer stolen information \cite{mitre_t1567}. In agentic systems, similar risks may arise when an agent retrieves confidential information and subsequently sends it outward through communication tools such as email or network APIs. Such workflows can cause private data exfiltration and may further lead to multi-step information leakage across the execution chain \cite{dhodapkar2026safetydrift}.

\subsubsection{File System Enumeration}
Via file system enumeration, an attacker or an attacker-guided agent can discover files, directories, and storage locations to identify sensitive information and support subsequent actions such as targeted collection, exfiltration, or destructive operations.  
In OpenClaw agents, file-search and local-execution capabilities make such discovery processes directly actionable during task execution.  
For instance, APT28 used the \textit{Forfiles} utility to locate PDF, Excel, and Word documents during data collection. 
Similarly, InvisibleFerret identified specific directories and files for exfiltration through the \textit{ssh\_upload} subcommands \textit{sdir} and \textit{sfind}, together with native search commands on both Windows and macOS platforms~\cite{mitreT1083}.

\subsubsection{Context Leakage}
Context leakage refers to the unauthorized disclosure of private information contained in the agent's runtime context. Such information may include augmented contextual knowledge, tool outputs, intermediate retrieval results, or interaction records, and it may be exposed through model outputs or agent actions~\cite{owasp2024llm06}. In OpenClaw agents, this risk is amplified as contextual memory, tool outputs, and intermediate retrievals may be integrated into the subsequent reasoning pipeline. A representative instance is MEXTRA, which induces agents to reveal private historical queries stored in long-term memory \cite{wang-etal-2025-unveiling-privacy}. Another instance is prompt-injection-based data exfiltration in tool-calling agents, where malicious task content causes agents to expose sensitive data during tool execution \cite{alizadeh2025simple}.

\textit{Summary.} Overall, the above attack mechanisms indicate that tool misuse and exploitation in OpenClaw agents extend beyond explicitly malicious tools and can also arise through unsafe composition, biased selection, iterative misuse, and covert outward transmission. Despite their distinct manifestations, these attacks fundamentally blur the boundary between benign capability invocation and harmful execution outcomes. Therefore, execution-layer security requires not only detecting individual malicious actions, but also preventing benign tools from being combined, redirected, or reused in ways that violate the user’s original intent and the system’s security constraints.

\subsection{Agentic Supply Chain Vulnerabilities}
\begin{figure}[!t]
\centering
\setlength{\abovecaptionskip}{-0.cm}
\includegraphics[width=1.0\linewidth]{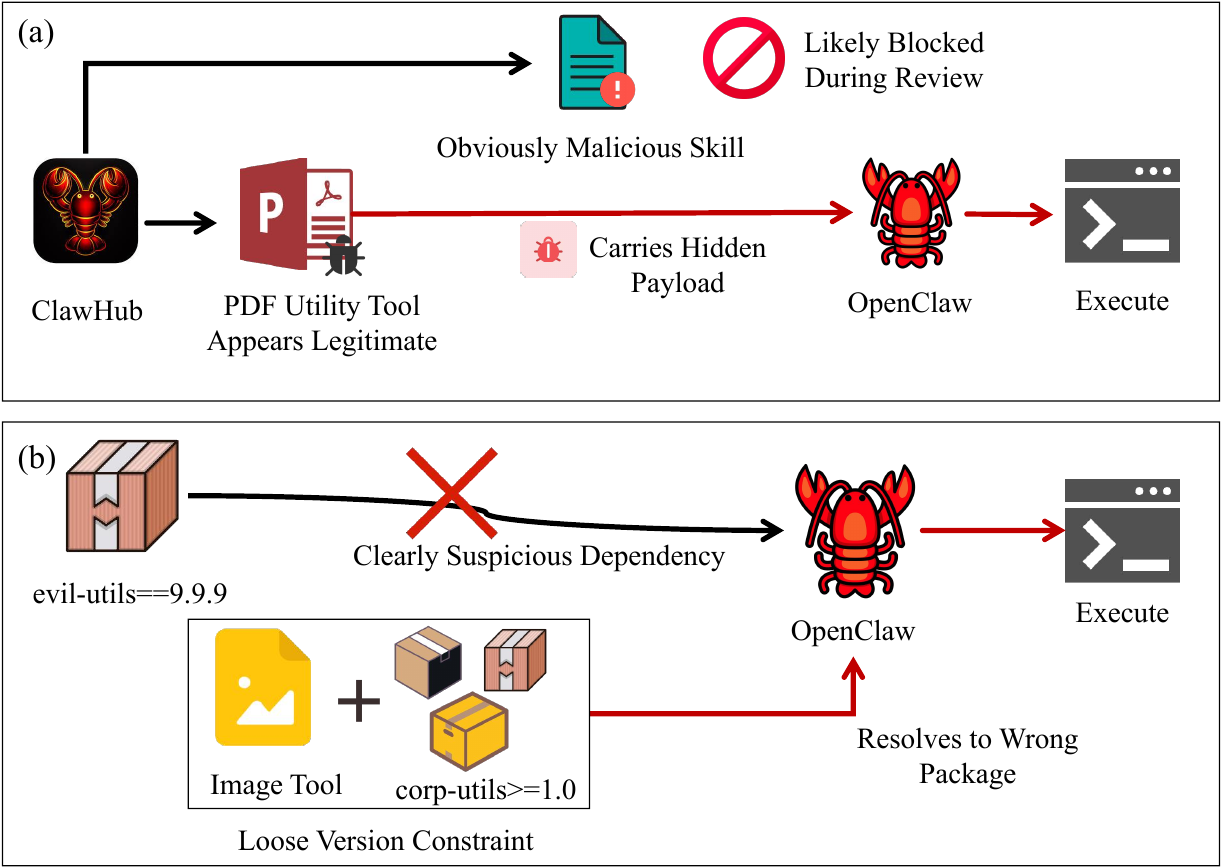}
\caption{Illustration of agentic supply chain vulnerabilities: (a) ClawHub poisoning, (b) unpinned dependencies.}
\label{fig:supplychain}\vspace{-3mm}
\end{figure}

Agentic supply chain vulnerabilities refer to a class of threats in which the trust assumptions underlying skill repositories, dependency resolution, and runtime resource acquisition are compromised during installation or execution, as shown in Fig. \ref{fig:supplychain}. Rather than directly manipulating the agent’s reasoning goals, these attacks target the external software and content supply channels on which OpenClaw depends. Although such skills, packages, or downloaded resources may initially appear legitimate, they can still introduce malicious logic into the execution pipeline. As OpenClaw agents can install, invoke, and combine third-party skills with local tools and scripts, such a compromise may further affect capability selection, dependency integrity, remote code retrieval, and execution transparency, thereby undermining the reliability and security of the overall agent system. In practice, these threats are carried out in various forms, including ClawHub poisoning, skill poisoning, capability impersonation, dependencies without fixed versions, external script fetching, and obfuscated code.

\subsubsection{ClawHub Poisoning}
In ClawHub poisoning attacks, adversaries usually publish malicious skills on the ClawHub marketplace or disguise harmful skills as benign ones. Once installed, these skills may introduce unsafe dependencies, hidden instructions, or attacker-controlled execution logic into the agent workflow~\cite{li2026secureagentskills,ying2026uncovering}. In OpenClaw agents, this threat is particularly severe because installed skills can be directly integrated into the execution pipeline and repeatedly invoked in later tasks. A representative example is the ClawHavoc campaign, in which Koi Security identified 341 malicious skills on ClawHub. Although these skills appeared to be ordinary utility tools, they contained harmful initialization steps that downloaded and executed harmful payloads~\cite{koi2026clawhavoc}. Similarly, the ToxicSkills study analyzed 3,984 skills and identified 76 malicious payloads, showing that poisoned skills may combine hidden prompt injection with credential theft, malware delivery, or remote instruction loading~\cite{beurerkellner2026toxicskills}.

\subsubsection{Skill Poisoning \& Capability Impersonation}
In this attack, adversaries can manipulate skill descriptions, configuration settings, or bundled model files, so that harmful skills appear to be legitimate and trustworthy ones \cite{tie2026badskill}. In OpenClaw agents, such attacks undermine the trust assumptions used when the agent selects and invokes skills during task execution. For instance, in the “hacked-weather” case, manipulated descriptions and priority settings caused ordinary weather-related queries to be routed to attacker-controlled code \cite{deng2026taming}. In a real-world incident reported by Trend Micro, malicious OpenClaw skills were disguised as required dependencies in configuration files such as \textit{SKILL.md}. As a result, the system installed the AMOS malware, leading to the exfiltration of documents and keychain data \cite{oliveira2026amos}.

\subsubsection{Unpinned Dependencies}
Unpinned dependencies arise when software components fail to fix dependency versions through lockfiles or similar constraints. As a result, dependency resolution may vary across environments, creating opportunities for version drift and malicious package substitution~\cite{sgtech2026pinning,kydyraliev2024depconfusion}. 
In OpenClaw agents, such instability may affect skills, tool wrappers, and script runtimes that are installed or resolved during runtime. 
A representative example is Birsan's dependency-confusion attack, in which internal package names used by organizations such as PayPal were registered on public package registries and subsequently resolved to attacker-controlled packages. Similarly, Python installations using ``--extra-index-url'' may prioritize higher-version public packages over private ones, which exposes enterprise software environments such as Microsoft's .NET Core to such risks~\cite{birsan2021dependencyconfusion}.

\subsubsection{External Script Fetching}
External script fetching means that the system retrieves and executes remotely hosted scripts or payloads during runtime. 
Such behavior can bypass local verification and static inspection, since external network content is directly transformed into executable state~\cite{liu2026agentskills,shan2026openclaw}. 
In OpenClaw agents, this risk arises when a skill setup process downloads remote code during installation or execution. For instance, TeamTNT used \textit{curl}, \textit{wget}, and \textit{batch} scripts to pull malicious tools from external sources after the initial compromise \cite{mitreG0139TeamTNT}. Similarly, the Water Curupira Pikabot distribution used \textit{Curl.exe} to download the Pikabot payload from remote servers into temporary directories, thereby directly preparing subsequent execution stages \cite{mitreT1105}.

\subsubsection{Obfuscated Code}
Obfuscated code refers to the use of encoding, encryption, packing, or similar transformations to conceal malicious semantics and evade detection \cite{mitreT1027}. In OpenClaw agents, such threats are often embedded within skill installation workflows to hide the actual execution behavior. For instance, Trend Micro reported malicious skills that decode Base64-encoded shell commands during installation to deploy the AMOS payload \cite{oliveira2026amos}. Similarly, VirusTotal identified seemingly benign skills that redirect execution to Base64-obfuscated scripts for malware retrieval, showing that obfuscated codes can conceal the true execution path within skill-installation workflows in OpenClaw agents \cite{quintero2026openclaw}.

\textit{Summary.} Overall, the above attack mechanisms indicate that agentic supply chain vulnerabilities in OpenClaw agents are not limited to obviously malicious skills, but can also emerge through disguised capabilities, unstable dependency resolution, remotely fetched code, and concealed execution logic. Despite their diverse forms, these threats erode the integrity of the software and content sources on which OpenClaw agents rely. As a result, untrusted components may enter the agent workflow and influence downstream planning and execution. Therefore, securing the OpenClaw supply chain requires not only inspecting skill functionality itself, but also strengthening repository governance, dependency pinning, provenance verification, and execution transparency across the entire lifecycle of agent supply chain.

\subsection{Unexpected Code Execution }
\begin{figure}[!t]
\centering
\setlength{\abovecaptionskip}{-0.cm}
\includegraphics[width=1.0\linewidth]{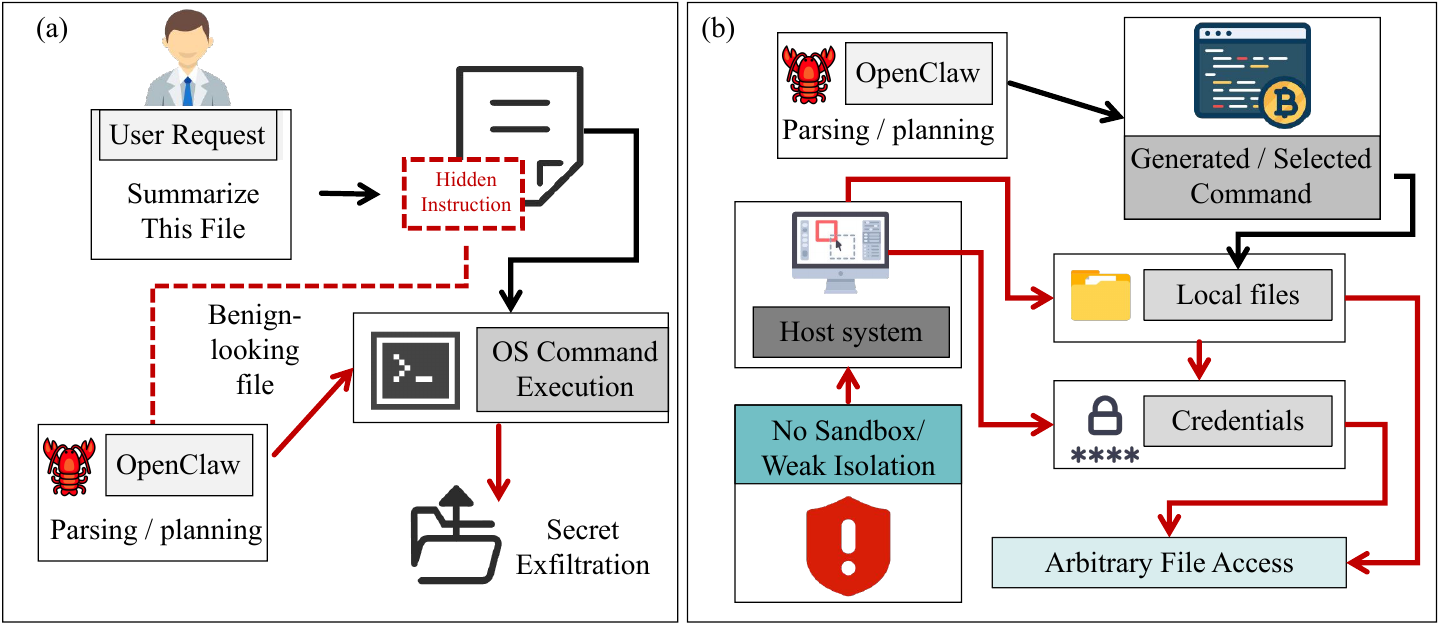}
\caption{Illustration of unexpected code execution: (a) OS command execution via prompt injection, (b) local execution without sandboxing.}
\label{figRCE}\vspace{-2mm}
\end{figure}

In OpenClaw agents, unexpected code execution is a class of threats in which natural-language inputs, intermediate outputs, or skill-related content are transformed into executable code, shell commands, or host-level actions beyond the user’s intended scope, as shown in Fig. \ref{figRCE}. Unlike common tool misuse, this threat can cross the boundary between semantic interpretation and system execution, allowing attacker-controlled content to influence command invocation, repair loops, output forwarding, or local execution contexts. As OpenClaw agents can translate high-level user instructions into executable operations such as shell commands, file modifications, and API calls, seemingly benign tasks such as summarization, installation, or error recovery may become entry points for OS-level compromise or persistent abuse. Representative forms of this threat include prompt injection that leads to OS command execution, uncontrolled agent-generated self-repair execution, unvalidated output forwarding to shell interfaces, and local execution without sandbox isolation.

\subsubsection{OS Command Execution Via Prompt Injection}
OS command execution via prompt injection occurs when attacker-controlled text is added into the agent's context as seemingly benign content, but is later interpreted as executable instructions. As a result, the agent may invoke OS-level execution interfaces beyond the user’s actual intent~\cite{debenedetti2024agentdojo}. 
In OpenClaw agents, this threat is particularly severe because natural-language instructions can be directly translated into shell commands, file operations, and other executable actions.  
For instance, prior analysis shows that hidden instructions embedded in a seemingly benign document can cause the agent, when merely asked to summarize the document, to export environment variables and upload them through \emph{curl}. This may leak sensitive data such as API keys or credentials~\cite{shan2026openclaw}. 
Another instance is Auto-GPT exploits, where adversarial web content induced the agent to generate and execute Python code during a normal summarization task. Such execution could further lead to host-level compromise after restart~\cite{euler2023autogptRCE}.

\subsubsection{Agent-Generated “Self-Repair” Runaway Execution}
Agent-generated “self-repair” runaway execution refers to a failure mode in which an agent, after encountering errors, autonomously enters a repair loop that continuously generates retry attempts, workaround commands, or additional tool calls without a clear stopping condition \cite{shi2025hafixagent}. 
This threat is particularly dangerous when the repair logic is influenced by adversarial feedback or persistent failure signals. Under such conditions, the agent may continue expanding its execution scope and consuming computational resources even though the original task is no longer making meaningful progress. 
In OpenClaw agents, Clawdrain shows that Trojanized skills can intentionally generate repair-related feedback to keep the agent trapped in repeated verification and repair cycles, thereby both tool invocation frequency and token consumption increase substantially \cite{dong2026clawdrain}.

\subsubsection{Unvalidated Output Forwarding to Shell}
Unvalidated output forwarding to shell happens when the text generated by the LLM or external tools is passed to a shell execution interface without sufficient validation, sanitization, or contextual filtering. As a result, untrusted outputs may be directly interpreted as executable OS commands. This behavior inherits the classical risk of OS command injection in agent systems~\cite{owasp2025llm02,mitre2024cwe78}. 
In OpenClaw agents, this threat is important as intermediate outputs from skill execution, web retrieval, or installation workflows may be implicitly trusted and forwarded to downstream shell-based actions. 
A representative real-world case is the AMOS campaign, in which malicious OpenClaw skills and related installation content induced the execution of encoded shell payloads that subsequently fetched and launched malware~\cite{oliveira2026amos}. 
Similarly, recent OpenClaw security analyses show that attacker-influenced content can be transformed into obfuscated shell-level payloads, indicating that unsafe forwarding of intermediate outputs can directly result in harmful execution~\cite{shan2026openclaw}.

\subsubsection{Local Execution Without Sandboxing}
Local execution without sandboxing refers to an execution design where code, commands, or skill actions produced by the agent are run directly on the host system without strong isolation boundaries. As such, untrusted model outputs can get the host’s file, process, and network privileges. Consequently, failures in validation or insufficient prompt control may escalate into direct system compromise rather than remaining confined to a single session \cite{irwin2025code}. This risk is obvious in OpenClaw agents, as they operate on user-controlled devices and can directly access local system resources \cite{wang2026agentassetrealworldsafety}. Structural weaknesses can further amplify this threat. For instance, failures in sandbox-policy enforcement and time-of-check-to-time-of-use (TOCTOU) flaws in path validation have shown that seemingly restricted sessions may still access sensitive host tools or escape workspace boundaries to perform arbitrary file read/write operations \cite{bors2026escaping}. Real-world incidents further demonstrate the practical impact of this threat. In the AMOS campaign, malicious OpenClaw skills can embed fake prerequisite instructions into bundled skill documentation, inducing users or systems to execute local Bash commands. This subsequently enabled the exfiltration of documents, credentials, and keychain data from the host system \cite{oliveira2026amos}.

\textit{Summary.} Overall, unexpected code execution in OpenClaw agents arises not only from unsafe model outputs, but also from the collapse of the boundary between language interpretation and system execution. As a result, seemingly benign inputs, repair signals, or intermediate outputs may be transformed into direct host-side actions when sufficient validation, isolation, or stopping controls are absent.

\subsection{Cascading Failures}
\begin{figure}[!t]
\centering \setlength{\abovecaptionskip}{-0.cm}
\includegraphics[width= 1.0\linewidth]{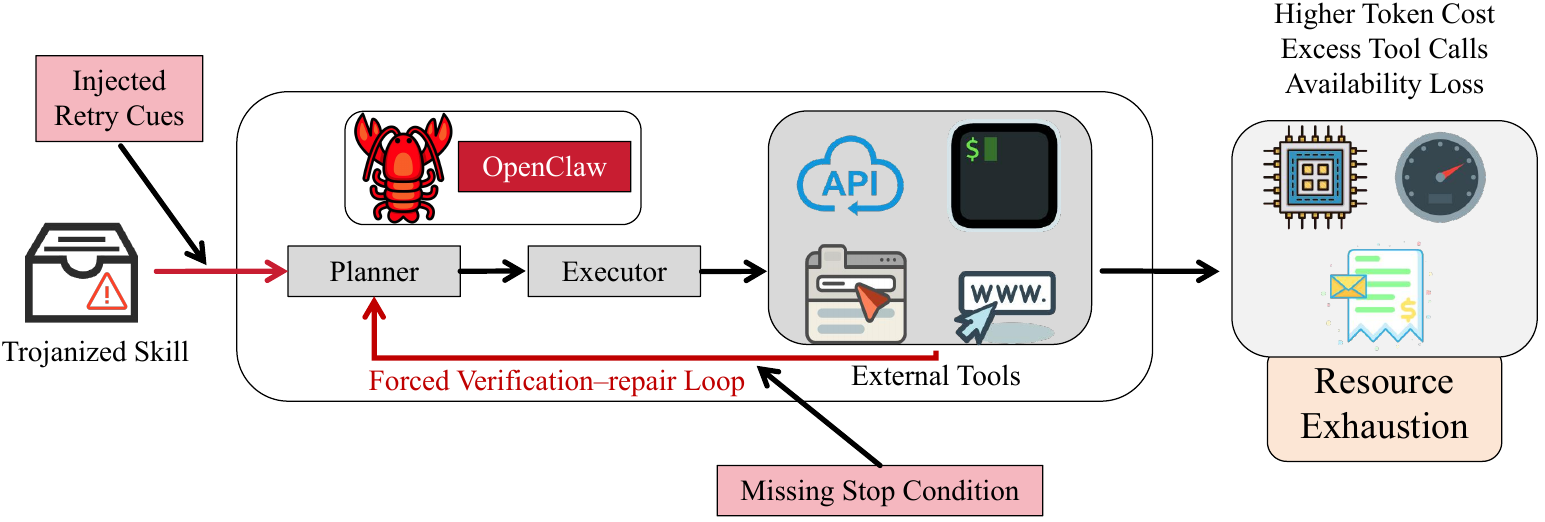}
\caption{Illustration of cascading failures: loop amplification \& resource overload threat.}\label{Cascading Failures}\vspace{-3mm}
\end{figure}

Cascading failures refer to a class of risks in which an initially local flaw, poisoned state, or adversarial deviation does not stay a single step. Instead, they move across multiple stages of the OpenClaw pipeline and become amplified during later reasoning and execution, as shown in Fig. \ref{Cascading Failures}. 
Different from isolated tool misuse or single execution errors, cascading failures can form a chain reaction across planning, memory, execution, and feedback loops. An early compromise may persist across stage transitions and subsequently trigger broader downstream effects. This risk is particularly significant in OpenClaw agents because they integrate persistent states, planner-executor cooperation, and external tool calls. Under such conditions, a small initial fault may gradually evolve into a durable system-level failure.  
In practice, cascading failures may emerge through cross-stage compound threats, fault persistence in saved state, planner-executor coupling fan-out, and loop amplification or resource overload.

\subsubsection{Cross-Stage Compound Threats}
Cross-stage compound threats arise when malicious influence introduced at one stage of the agent workflow persists into later stages and is further amplified there. As a result, the final damage is caused not by a single isolated flaw, but by the propagation and accumulation of compromise across multiple stages \cite{banerjee2026cascade}. In OpenClaw agents, this risk is particularly severe, as the agent maintains persistent capability state, identity state, and knowledge state across tasks, while also operating with broad local execution privileges. Existing evaluations show that, when attackers compromise any one of these state dimensions, the average success rate of attacks can rise substantially from 24.6\% to 64--74\% \cite{wang2026agentassetrealworldsafety}.

\subsubsection{Fault Persistence in Saved State}
Fault persistence in saved state refers to cases where attacker-influenced or unverified information is written into persistent state and later reused as trusted context. 
As a result, temporary errors may persist across sessions and continue to influence subsequent reasoning and actions~\cite{bousetouane2026memorycontrol,yang2026zombie}. 
For instance, as reported in~\cite{wang2026agentassetrealworldsafety}, knowledge poisoning can cause OpenClaw agents to retain fabricated refund preferences in long-term memory. Under a later and seemingly benign request, this poisoned memory can trigger unauthorized batch refunds. 
Similarly, false policy rules injected into persistent memory can alter the OpenClaw agent’s future decisions in later sessions, transforming temporary input manipulation into long-term behavioral distortion~\cite{deng2026taming}.

\subsubsection{Planner-Executor Coupling Fan-out}
Planner-executor coupling fan-out refers to situations where a flawed or adversarially influenced planning decision is expanded by executors into multiple downstream actions. 
As a result, the risk grows through action fan-out instead of remaining confined to a single local error~\cite{shi2025progent,dhodapkar2026safetydrift}. 
In OpenClaw agents, this threat can be amplified as the agent can directly invoke external tools and execute OS-level operations~\cite{ying2026uncovering}. 
For instance, in the Ash's owner-spoofing incident, one incorrect authority judgment cascaded into multiple high-impact actions, including workspace overwrite and admin reassignment. In the corrupted-constitution incident, an externally editable governance document caused the agent to attempt peer-agent shutdowns, remove users, and propagate compromised policies across other agents~\cite{shapira2026agentschaos}.

\subsubsection{Loop Amplification \& Resource Overload}
Loop amplification and resource overload occur when adversarial inputs, cross-agent interactions, or malicious skill logic cause an agent to repeatedly perform actions or continuously invoke costly tools. As a result, excessive consumption of tokens, computational resources, storage, or network bandwidth may occur, ultimately degrading system availability~\cite{owasp2025unbounded}. 
In OpenClaw agents, the \emph{Agents of Chaos} study shows that a non-owner could trap two agents in a circular interaction for at least nine days, consuming about 60,000 tokens. The same study also shows that an attacker could induce an agent to create a background polling job without a clear stopping condition~\cite{shapira2026agentschaos}. 
Another OpenClaw study, \emph{Clawdrain}, shows that a Trojanized skill can increase token usage by about six to seven times compared with benign execution, and by about nine times in a more costly failure setting~\cite{dong2026clawdrain}.

\textit{Summary.} Overall, cascading failures represent a systemic amplification process in which initially bounded faults propagate across stages and evolve into broader behavioral, operational, or availability damage. Rather than viewing these incidents as isolated failures in memory, planning, or execution, they should be regarded as an interconnected risk pattern characterized by persistence, propagation, and amplification. This phenomenon highlights how tightly coupled OpenClaw agent pipelines can transform localized vulnerabilities into large-scale failures during long-horizon autonomous operation.

\subsection{Existing/Potential Countermeasures}
The execution-layer threats discussed above can be mitigated through the following complementary defensive mechanisms.

\subsubsection{Malicious Tool-Use Chain Signature Library}
As reported in \cite{li2025stac}, seemingly benign tool invocations can be chained into harmful execution flows, highlighting the necessity of behavior-sequence-aware detection mechanisms.
Similar to the design of virus signature databases, agentic systems can maintain a signature library that captures behavioral patterns (e.g., tool invocation order, data-flow dependencies, and cross-step interactions) of known malicious tool-use chains \cite{hu2026maltool}. During execution, a monitoring module could compare the current tool invocation trace and data flow against recorded signatures. If a high-confidence match is detected, the system can trigger alerts, suspend further execution, or require additional verification before proceeding with the tool chain. 

\subsubsection{Supply Chain Verification}

To mitigate supply chain risks, skills, scripts, dependencies, and remotely fetched resources should be verified before installation or execution \cite{zhao2025mind}. 
The system should not rely only on package metadata or repository origin. Instead, it should validate the provenance and integrity of external components, including dependency versions, installation scripts, remote resources, and other code-related metadata \cite{10.1145/3704724}. 
If suspicious dependencies, scripts, or code are identified, the installation or execution process should be restricted until further verification is completed.

\subsubsection{Sandbox Execution and Validation}

Sandbox execution is mainly used to isolate operations that may contain potential risks by running them inside restricted execution environments.  
In agentic systems, shell commands, file operations, and code execution should be confined through permission and resource constraints, ensuring that these actions can only access the resources required for the current task while preventing unnecessary or unauthorized access.
Before execution, the system should further verify whether the actions remain consistent with the original task objective and whether they exceed the predefined permission scope.

\section{Interaction-Layer Threats \& 
Countermeasures of OpenClaw Agents}\label{SECURITY3}

Interaction-layer threats in OpenClaw agents refer to security risks that arise from the agent's continuous interactions with users, external services, peer agents, and the underlying system environment. These threats may directly affect authentication mechanisms, privilege boundaries, communication channels, and user approval workflows. As OpenClaw agents become increasingly integrated with email, calendars, file systems, web APIs, and local OS tools, interaction can no longer be regarded as a simple exchange of information. Instead, it becomes a critical control surface through which identity misuse, privilege escalation, and system compromise may occur. This section examines interaction-layer threats faced by OpenClaw agents and discusses corresponding mitigation strategies.

\subsection{Identity and Privilege Abuse}
\begin{figure}[!t]
\centering \setlength{\abovecaptionskip}{-0.cm}
\includegraphics[width= 1.0\linewidth]{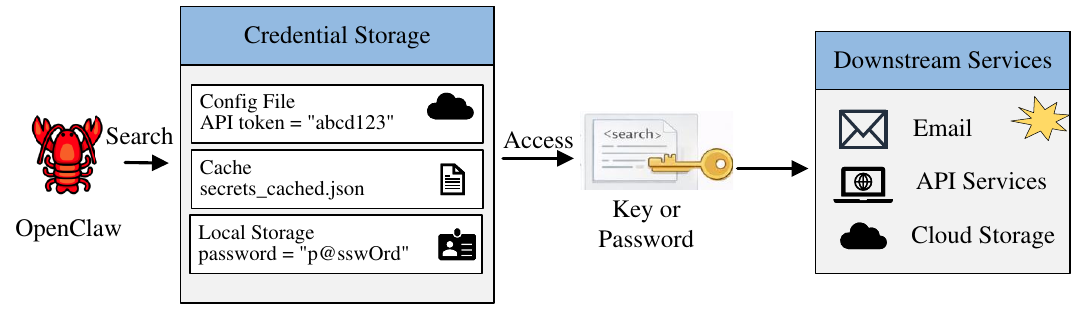}
\caption{Illustration of identity and privilege abuse: credential access.}\label{figco1}\vspace{-3.5mm}
\end{figure}

Identity and privilege abuse refers to a class of threats in which adversaries exploit weaknesses in authentication, authorization, or credential management to obtain unauthorized access or elevated execution capabilities, as shown in Fig.~\ref{figco1}. These attacks may include credential theft, abuse of authorization channels, or extraction of sensitive configuration artifacts. Once successful, they can allow the agent to act beyond its intended permission scope.
Once vulnerabilities exist in identity verification, token handling, or privilege assignment, even limited interaction capabilities can be escalated into high-impact control over system resources. As a result, local misuse at the interaction layer may evolve into broader system compromise through privilege amplification. In practice, identity and privilege abuse may arise through several representative mechanisms, including gateway authentication misconfiguration, unauthorized credential access, and excessive permission requests. 

\subsubsection{Gateway Authentication Misconfiguration}
Gateway authentication misconfiguration refers to improper handling of gateway connection parameters that can result in credential leakage and subsequent privilege escalation. In particular, flaws in WebSocket gateway configuration may allow attacker-controlled endpoints to receive sensitive authentication tokens during connection establishment. For instance, when the control interface trusts the \textit{gatewayUrl} parameter from the query string and automatically establishes a connection during page initialization without proper validation, the browser may send the stored gateway token to an attacker-controlled server. Then, the adversary can reuse the leaked token to authenticate with the local gateway, modify sandbox or tool policies, and invoke privileged operations. In severe cases, this may lead to remote code execution (RCE). For OpenClaw agents, the risk lies not only in token theft, but also in the close coupling between gateway authentication and execution control.

\subsubsection{Unauthorized Credential Access}
Unauthorized credential access refers to situations in which attackers acquire or reuse authentication credentials, such as API keys, that have not been properly protected. This issue is particularly relevant in the context of OpenClaw agents, as they often need to load such credentials to access email services, cloud APIs, and local tools. These credentials may be stored in configuration files, cache directories, or remain temporarily in process memory. If such credentials are exposed, especially when they are long-lived or stored in plain text, attackers may be able to reuse them outside their intended environment to gain unauthorized access. Consequently, credential leakage is not merely a local security concern, but can also serve as a starting point for broader attacks. This threat pattern aligns with credential access techniques described in the ATT\&CK framework, where exposed secrets are collected and used for subsequent intrusion activities.

\subsubsection{Excessive Permission Requests}
Excessive permission requests occur when OpenClaw agents, plugins, or third-party services request privileges that far exceed what is necessary for completing the current task. The risks associated with such over-privileged access are not confined to the initial authorization stage. Once granted by the user, these permissions may remain valid across multiple sessions and may be invoked in later operations that the user does not anticipate. In OAuth-based authorization environments, the risks of long-term retained permissions are further amplified due to the persistence and reusability of granted scopes. Malicious applications may abuse user consent, elevated roles, or application-level credentials to obtain and retain high-level access privileges \cite{mahara2025oauth}.

\textit{Summary.} Identity and privilege abuse is a critical risk, as it may enable an OpenClaw agent to operate with authority beyond its intended scope. This can occur through leaked tokens or service credentials, as well as through overly broad permission assignments. Once exposed, these access channels may be reused by attackers to perform actions that extend beyond the original task context. Moreover, such permissions may remain valid beyond the initial interaction, persisting into subsequent sessions and influencing later tool invocations or service connections. As a result, interaction-layer weaknesses may spread into the runtime environment and expose deeper system components to unauthorized access or control.

\subsection{Insecure Inter-Agent Communications}

\begin{figure}[!t]
\centering \setlength{\abovecaptionskip}{-0.cm}
\includegraphics[width= 0.78\linewidth]{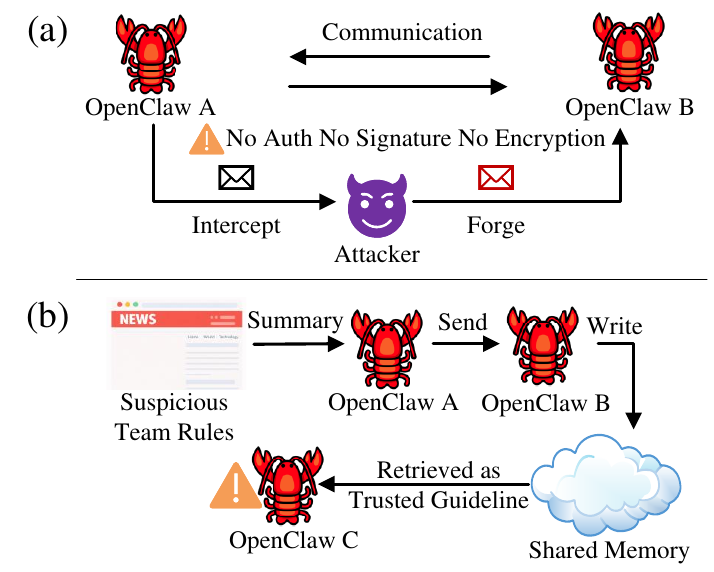}
\caption{Illustration of insecure inter-agent communication: (a) CIA issues in inter-agent communication, (b) cross-context contamination.}\label{figco2}\vspace{-2mm}
\end{figure}

As shown in Fig. \ref{figco2}, security risks may arise during communication among collaborative agents. These risks typically arise when the system fails to verify the identity of the message sender, ensure message integrity, or correctly establish the execution context of an instruction. For instance, an agent may mistakenly execute outdated instructions or mix information from different interaction contexts. Such insufficient verification can further lead to instruction contamination, identity confusion, misplaced trust, and the propagation of malicious payloads. More critically, these effects are not necessarily confined to the initially receiving agent. Once a message is accepted as trustworthy, it may be summarized, reused, or relayed in subsequent interactions, thereby influencing downstream agents in the collaboration chain. In OpenClaw agents, this issue is exacerbated by the use of heterogeneous communication interfaces, such as email and webhooks. In the following, we discuss how weak confidentiality, integrity, and authentication (CIA) guarantees, cross-context contamination, trust-chain replay, protocol downgrades, and descriptor forgery collectively expand the attack surface of inter-agent communication.

\subsubsection{CIA Issues in Agent Communications}
When agents are unable to reliably verify message provenance, ensure content integrity, or enforce appropriate access control for recipients, inter-agent communication becomes inherently fragile. These weaknesses commonly stem from insufficient authentication, weak or missing integrity verification, and inadequate protection of confidential information. In multi-agent systems, a falsified or tampered message is not necessarily contained at the point of reception. Once accepted by one agent, it may be reused, summarized, or propagated to other agents, thereby influencing subsequent decisions across the collaboration process and eventually turning into much broader risks. 

\subsubsection{Cross-Context Contamination}
Cross-context contamination occurs when a tampered or adversarially crafted message exceeds its original semantic or operational scope and becomes embedded in another agent’s reasoning, memory, or action processes. This risk typically emerges when messages are modified, injected with malicious content, or reformatted during transmission and forwarding, thereby altering their meaning or intent. In multi-agent systems, such contaminated information can propagate further through shared communication channels, summarization pipelines, or persistent memory mechanisms, resulting in cascading effects across multiple agents. Shapira et al. \cite{shapira2026agentschaos} show that interactions among multiple agents can amplify failures originating in a single agent and allow security vulnerabilities to propagate alongside agent capabilities and permissions. For OpenClaw agents, the threat extends beyond simple content tampering to include the persistence of contaminated information across contexts and its potential influence on future interactions.

\subsubsection{Replay Attacks on Trust Chains}

Replay attacks in trust chains occur when an attacker submits an authorization message, token exchange, or trust confirmation that was legitimately issued in a previous interaction. 
The core issue lies not in the fabrication of invalid credentials, but the reuse of valid ones outside their intended temporal or contextual scope. As a result, conventional authenticity checks may fail to detect such attacks, since the credential itself originates from a trusted source. This risk is significant to OpenClaw agents, where inter-agent trust may depend on cached approvals, prior interaction histories, or time-limited tokens. Without mechanisms for freshness validation and context binding, outdated authorizations can be incorrectly interpreted as current, effectively reconstructing trust relationships that are no longer valid. Similar replay-related risks are also highlighted in OAuth 2.0 threat models, which recommend mitigations such as short-lived tokens, single-use credentials, and audience-bound token design to reduce the likelihood of unauthorized reuse.

\subsubsection{Protocol Downgrade \& Descriptor Forgery}

Protocol downgrade and descriptor forgery target the cues that agents rely on prior to evaluating message content. In such attacks, an adversary may push communication into a less secure protocol configuration or manipulate metadata fields such as sender identity, role labels, or other contextual attributes associated with the message. As a result, the receiving agent may misinterpret the origin, status, or authority level of the message. This is especially dangerous for OpenClaw agents, since their trust judgments are derived not only from visible text but also from associated descriptors to that text. Once these metadata fields are falsified or altered, malicious input may be incorrectly treated as originating from a legitimate and trusted source. This can cause the agent to assign inappropriate levels of trust before examining the actual content.

\textit{Summary.} Security threats in inter-agent communication are not simple transmission errors or misinterpretations. They instead reflect fundamental limitations in the system’s trust management mechanisms, including how agents verify each other’s authenticity, assess the reliability of shared information, and coordinate actions based on exchanged messages. Each of these components introduces potential vulnerabilities. Once a single agent is compromised, such weaknesses can propagate rapidly through the system via message forwarding, shared memory, or collaborative decision-making processes. Over time, this can lead to deviations from intended behavior or even loss of control over collective task execution within the multi-agent system.

\subsection{Human-Agent Trust Exploitation}

\begin{figure}[!t]
\centering \setlength{\abovecaptionskip}{-0.cm}
\includegraphics[width= 1.0\linewidth]{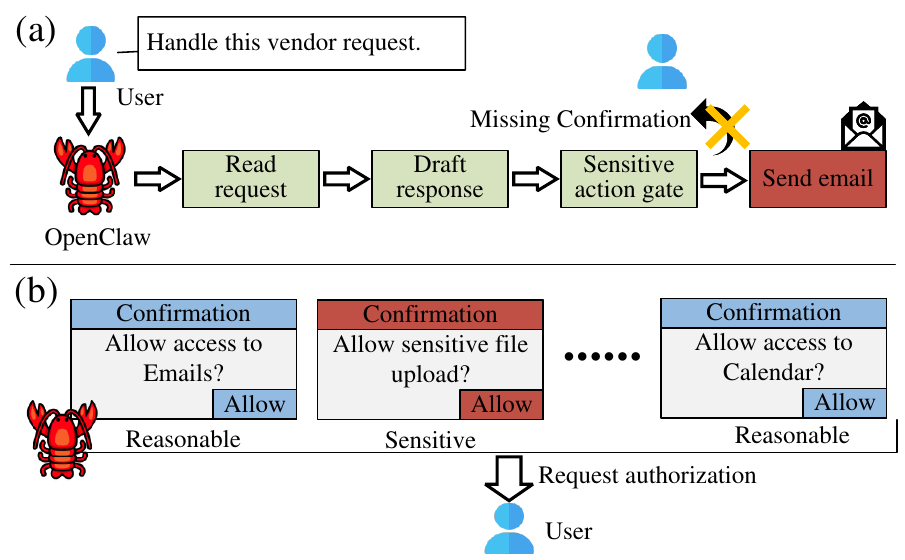}
\caption{Illustration of human-agent trust exploitation threats: (a) missing confirmation for sensitive actions, (b) consent fatigue.}\label{figLLM3}\vspace{-2mm}
\end{figure}

Human-agent trust exploitation refers to threats in which adversaries leverage user's default trust in an OpenClaw agent's capability. By manipulating interaction design, authorization flows, or feedback mechanisms, attackers can influence how users judge risks and make decisions, as shown in Fig.~\ref{figLLM3}. Unlike attacks targeting the agent's internal reasoning, this threat mainly manifests at the human-agent interaction layer. In this context, users may lower their level of scrutiny, misinterpret authorization boundaries, or approve actions that warrant closer verification. As OpenClaw agents increasingly support information filtering, recommendation, and autonomous execution tasks, users may be more inclined to treat their outputs as reliable inputs for decision-making, thereby amplifying the associated risks. Representative forms of this threat include omission of confirmations for sensitive operations and consent fatigue.

\subsubsection{Missing Confirmation for Sensitive Actions}
The absence of confirmation mechanisms for sensitive operations occurs when an OpenClaw agent performs high-risk actions without obtaining explicit user authorization or secondary verification. As highlighted by OWASP, agents without proper review and authorization controls may act on ambiguous instructions or make flawed judgments. This can lead to severe outcomes, including data corruption or disruption of the user’s operating environment. The risk becomes more significant when OpenClaw agents are integrated with high-privilege resources, such as email, payment services, or device management interfaces. In such settings, a single vague instruction or a one-time permission grant may be incorrectly interpreted as standing authorization for subsequent sensitive actions. The agent may then delete files, send messages in bulk, or modify system settings without further confirmation. At that point, the agent is no longer merely assisting the user. Instead, it makes critical decisions on the user’s behalf, even in situations that require explicit consent. This undermines user control and increases the likelihood of security incidents.

\subsubsection{Consent Fatigue}

Consent fatigue refers to the phenomenon in which users are repeatedly exposed to authorization prompts. Over time, they may stop examining the specific content carefully and instead click “agree” out of habit. This issue becomes more serious when pop-up windows use similar visual layouts and wording, or when the system fails to explain why a particular operation requires elevated attention. In such cases, user consent may shift from a deliberate judgment to a mechanical reaction. Cui et al. \cite{CHOI201842} link this phenomenon with privacy fatigue, arguing that repeated authorization requests reduce users’ willingness to inspect permission details thoroughly. As a result, users may grant permissions after only a quick glance. For OpenClaw agents, they frequently require user confirmation before tool calls, permission upgrades, intermediate operations, or runtime status changes. Although these prompts are intended to enhance security through user verification, their effectiveness can diminish when they appear excessively often and lack sufficient differentiation. After repeated exposure to similar prompts, users may begin approving requests reflexively by habitually clicking “agree”.


\textit{Summary.} Human-agent trust exploitation targets the user, who ultimately remains responsible for reviewing, approving, or acting upon the agent’s outputs. Adversaries may leverage users' trust in the agent's output to influence their judgment and acceptance of generated content. In OpenClaw agents, missing confirmations and repetitive authorization prompts can both weaken user control and increase security risks.

\subsection{Existing/Potential Countermeasures}
The above interaction-layer threats can be mitigated through the following defensive mechanisms.


\subsubsection{Agent Least Privilege}

To enforce the principle of least privilege for agents, authorization artifacts (e.g., access tokens, session credentials, and approval grants) should be bound to a specific task context rather than treated as persistent permissions \cite{zhao2025mind}. Each authorization artifact should have a clearly defined scope and expiration period, and every access request should be validated against the context of the original task \cite{zhao2026clawguard}. For instance, a short-lived token issued for reading files in task T1 should not be reusable in a different task T2.  If subsequent tasks require similar privileges, the agent must explicitly request new authorization.

\subsubsection{Secondary Verification for Security-Sensitive Actions}
To mitigate risks arising from human-agent trust exploitation, agents can employ secondary verification mechanisms before executing security-sensitive operations. Ambiguous or high-level user instructions should not be interpreted as implicit authorization for subsequent critical actions.
For sensitive operations such as deleting files, modifying system configurations, transferring data, or invoking high-privilege tools, the agent should explicitly request additional user confirmation before execution.

The following Table~\ref{tab:openclawsecuritysummary} summarizes representative literature on security threats across cognition, execution, and interaction levels, as well as potential countermeasures to OpenClaw agents.

\begin{table*}[ht]
    \centering
    \setlength{\abovecaptionskip}{0.1cm}
    \caption{Summary of Representative Literature on Security Threats and Potential Countermeasures to OpenClaw Agents}
    \label{tab:openclawsecuritysummary}
    \begin{tabular}{cclc}
    \hline
    \textbf{Ref.} & 
    \textbf{\begin{tabular}[c]{@{}c@{}}Security\\ Threat\end{tabular}} & 
    \multicolumn{1}{c}{\textbf{\begin{tabular}[c]{@{}l@{}}
    $\star$ Purpose\\
    $\bullet$ Advantages\\
    $\circ$ Limitations\\ 
    $\dagger$ Evaluation Metrics
    \end{tabular}}} & 
    \textbf{\begin{tabular}[c]{@{}c@{}}Potential\\ Defense Methods\end{tabular}} \\
    \hline

    \cite{2026arXiv260216958D} & 
    {{\begin{tabular}[l]{@{}c@{}} Structured Instruction\\ Injection \end{tabular}}} & 
    {\begin{tabular}[l]{@{}l@{}}
    $\star$ Automate agent hijacking via structural template injection \\
    $\bullet$ Targets role-boundary confusion and improves transferability across closed-source agents \\
    $\circ$ Effectiveness depends on chat-template parsing and context serialization assumptions \\
    $\dagger$ Attack success rate (ASR), query efficiency, utility score, confirmed vulnerability count
    \end{tabular}} & 
    \begin{tabular}[c]{@{}c@{}}Boundary \\isolation \end{tabular} \\
    \hline

    \cite{2026arXiv260105504D} & 
    {{\begin{tabular}[l]{@{}c@{}} Persistent Memory\\ Rule Injection \end{tabular}}} & 
    {\begin{tabular}[l]{@{}l@{}}
    $\star$ Evaluate memory poisoning attacks and defenses for memory-based LLM agents \\
    $\bullet$ Studies realistic memory states, retrieval settings, and trust-aware filtering \\
    $\circ$ Domain-specific evaluation; broader validation remains needed \\
    $\dagger$ ASR, memory append/rejection behavior, benign accuracy, false-positive rejection rate
    \end{tabular}} & 
    \begin{tabular}[c]{@{}c@{}} Memory \\sanitization\end{tabular} \\
    \hline

    \cite{2025arXiv250217832H} & 
    {{\begin{tabular}[l]{@{}c@{}} RAG Knowledge\\ Source Poisoning \end{tabular}}} & 
    {\begin{tabular}[l]{@{}l@{}}
    $\star$ Study RAG poisoning through local and global adversarial knowledge injection \\
    $\bullet$ Covers retriever, reranker, and generator stages in multimodal RAG pipelines \\
    $\circ$ Assumes attackers can insert poisoned entries into external knowledge bases \\
    $\dagger$ ASR, retrieval recall, answer accuracy
    \end{tabular}} & 
    \begin{tabular}[c]{@{}c@{}}Context \\Validation\end{tabular} \\
    \hline

    \cite{ying2026uncovering} & 
    {{\begin{tabular}[l]{@{}c@{}} Instruction Amnesia\\ via Context\\ Compression \end{tabular}}} & 
    {\begin{tabular}[l]{@{}l@{}}
    $\star$ Analyze OpenClaw threats across cognition, execution, and information-system dimensions \\
    $\bullet$ Proposes full-lifecycle agent security architecture \\with zero-trust execution, intent verification, and cross-layer correlation \\
    $\circ$ Mainly a defense blueprint; full deployment effectiveness needs broader validation \\
    $\dagger$ Threat taxonomy, attack-surface analysis, defense-layer design
    \end{tabular}} & 
    \begin{tabular}[c]{@{}c@{}}Intent \\Verification \end{tabular} \\
    \hline

    \cite{shi2026prompt} & 
    {{\begin{tabular}[l]{@{}c@{}} Tool Selection\\ Manipulation \end{tabular}}} & 
    {\begin{tabular}[l]{@{}l@{}}
    $\star$ Attack LLM-agent tool selection by injecting malicious tool documents into the tool library \\
    $\bullet$ Works in no-box settings and is evaluated across multiple LLMs and retrievers \\
    $\circ$ Requires the malicious tool document to enter the candidate tool library \\
    $\dagger$ ASR, attack hit rate, detector false-positive rate
    \end{tabular}} & 
    \begin{tabular}[c]{@{}c@{}}Tool-document \\vetting \end{tabular} \\
    \hline

    \cite{tie2026badskill} & 
    {{\begin{tabular}[l]{@{}c@{}} Skill Poisoning \&\\ Capability\\ Impersonation \end{tabular}}} & 
    {\begin{tabular}[l]{@{}l@{}}
    $\star$ Formalize backdoor attacks hidden inside model-bearing agent skills \\
    $\bullet$ Evaluates multiple skills and model architectures with high reported ASR \\
    $\circ$ Uses an OpenClaw-inspired simulation rather than a full marketplace deployment \\
    $\dagger$ ASR, benign-side accuracy, poison rate, perturbation robustness
    \end{tabular}} & 
    \begin{tabular}[c]{@{}c@{}}Provenance \\verification \end{tabular} \\
    \hline

    \cite{liu2026agentskills} & 
    {{\begin{tabular}[l]{@{}c@{}} External Script\\ Fetching \end{tabular}}} & 
    {\begin{tabular}[l]{@{}l@{}}
    $\star$ Conduct large-scale empirical analysis of real-world agent skill vulnerabilities \\
    $\bullet$ Builds a vulnerability taxonomy and a detection framework for agent skills \\
    $\circ$ Static analysis may mix malicious skills, insecure practices, and dual-use security tools \\
    $\dagger$ Vulnerability prevalence, precision, recall, odds ratio, severity distribution
    \end{tabular}} & 
    \begin{tabular}[c]{@{}c@{}}Repository \\governance\end{tabular} \\
    \hline

    \cite{debenedetti2024agentdojo} & 
    {{\begin{tabular}[l]{@{}c@{}} OS\\ Command Execution\\ Via Prompt Injection \end{tabular}}} & 
    {\begin{tabular}[l]{@{}l@{}}
    $\star$ Build AgentDojo to evaluate prompt injection attacks and defenses in dynamic tool environments \\
    $\bullet$ Provides realistic tasks and security test cases with environment-state-based checks \\
    $\circ$ Existing attacks and defenses remain limited; tool filtering can reduce utility \\
    $\dagger$ Utility, targeted ASR, utility under attack, defense success trends
    \end{tabular}} & 
    \begin{tabular}[c]{@{}c@{}}Tool\\ filtering \end{tabular} \\
    \hline

    \cite{shan2026openclaw} & 
    {{\begin{tabular}[l]{@{}c@{}} Unvalidated Output\\ Forwarding to Shell \end{tabular}}} & 
    {\begin{tabular}[l]{@{}l@{}}
    $\star$ Evaluate OpenClaw resilience against adversarial scenarios and propose a \\human-in-the-loop (HITL) defense layer \\
    $\bullet$ Covers multiple attack categories and improves defense over native OpenClaw behavior \\
    $\circ$ Pattern-based rules still have gaps against encoded payloads and sandbox escape cases \\
    $\dagger$ Defense rate, blocked severe attacks, category-wise defense rates
    \end{tabular}} & 
    \begin{tabular}[c]{@{}c@{}} HITL\\ Approval\end{tabular} \\
    \hline

    \cite{CHOI201842} & 
    {{\begin{tabular}[l]{@{}c@{}} Consent\\ Fatigue \end{tabular}}} & 
    {\begin{tabular}[l]{@{}l@{}}
    $\star$ Conceptualize privacy fatigue and examine its impact on online privacy behavior \\
    $\bullet$ Provides empirical evidence for fatigue-related weakening of privacy decision-making \\
    $\circ$ Not an agent-security paper; mainly supports the human-factor basis of consent fatigue \\
    $\dagger$ Path coefficient, coefficient of determination ($R^2$), effect size, composite reliability (CR),\\ average variance extracted (AVE)
    \end{tabular}} & 
    \begin{tabular}[c]{@{}c@{}}Secondary \\verification \end{tabular} \\
    \hline

    \end{tabular}
\end{table*}

\section{Future Research Directions}\label{sec:FUTUREWORK}
This section discusses several key research directions for advancing the security of OpenClaw-style agents.

\subsection{Security Benchmarking and Formalization}
Most existing studies on OpenClaw security focus on specific attack vectors or isolated system components, whereas security risks across the entire agent lifecycle remain insufficiently explored. Moreover, current evaluations primarily depend on manually constructed attack scenarios, making it challenging to compare different defense methods.

Future work should focus on developing unified security models that capture the full lifecycle of OpenClaw agents. One promising direction is the formalization of agent states, execution flows, and interaction dependencies, allowing security-relevant transitions during reasoning, planning, and execution to be represented in a structured and analyzable manner.
In addition, the development of standardized security benchmarks and publicly available attack datasets remains an important research direction. 

\subsection{Embodied Agent Security}
As OpenClaw agents increasingly interact with physical environments, their security risks are no longer limited to the digital domains. In embodied settings, the decisions made by the agents directly influence physical devices, infrastructure, and human users, making potential failures significantly more consequential.

This shift necessitates a re-evaluation of traditional security assumptions. The objective is no longer limited to preserving data integrity or ensuring correct execution, but also includes guaranteeing that physical actions remain safe and constrained. Given the tight coupling between reasoning processes and physical actuation in embodied agents, runtime monitoring and control mechanisms become particularly critical in safety-sensitive applications.

\subsection{Adaptive and Self-Evolving Security Mechanisms}
OpenClaw agents are typically deployed in long-running environments, where both system behavior and external conditions may evolve over time. In such settings, static security policies may not always remain effective, particularly when new attack patterns arise or when the boundaries of system interaction shift.

Future work should investigate adaptive security mechanisms that can co-evolve with the agent. One promising direction is the development of behavioral governance models driven by runtime observations, such as tool usage patterns, execution traces, and interaction histories. Leveraging these observations, security mechanisms can be gradually refined and updated, enabling more context-aware enforcement rather than relying exclusively on predefined static rules.

\section{Conclusions}\label{sec:CONSLUSION}
OpenClaw agents represent an emerging class of LLM-based autonomous systems. This survey systematically investigated evolving attack vectors and corresponding defense strategies for OpenClaw agents. We first analyzed the core architecture and unique features of OpenClaw agents. Then, we examined security vulnerabilities across the cognition, execution, and interaction layers. Representative defense methods and potential countermeasures were also reviewed. Finally, several open challenges and future directions in this domain were outlined. This work aims to inspire further research toward the secure design, deployment, and governance of OpenClaw-style agents and autonomous agent platforms.

\bibliographystyle{ieeetr} 

\bibliography{ref.bib}

@misc{OpenClawAgentTakeover,
  author = {Oasis Security},
  title = {OpenClaw Vulnerability: Website-to-Local Agent Takeover},
year = {2026},
  howpublished = {\url {https://www.oasis.security/blog/openclaw-vulnerability}},
  note = {Accessed on 2026-03-10}
}

@misc{CiscoOpenClaw,
  author = {Cisco},
  title = {Personal {AI} Agents like OpenClaw Are a Security Nightmare},
year = {2026},
  howpublished = {\url {https://blogs.cisco.com/ai/personal-ai-agents-like-openclaw-are-a-security-nightmare}},
  note = {Accessed on 2026-03-10}
}

@misc{OpenClawWebsite,
  author = {},
  title = {OpenClaw},
year = {2026},
  howpublished = {\url {https://openclaw.ai/}},
  note = {Accessed on 2026-03-15}
}

@misc{OpenClawGithub,
  author = {OpenClaw},
  title = {OpenClaw: Personal {AI} Assistant},
year = {2026},
  howpublished = {\url {https://github.com/openclaw/openclaw}},
  note = {Accessed on 2026-03-10}
}

@article{ying2026uncovering,
  title={Uncovering Security Threats and Architecting Defenses in Autonomous Agents: A Case Study of OpenClaw},
  author={Ying, Zonghao and Yang, Xiao and Wu, Siyang and Song, Yumeng and Qu, Yang and Li, Hainan and Li, Tianlin and Wang, Jiakai and Liu, Aishan and Liu, Xianglong},
  journal={arXiv preprint arXiv:2603.12644},
  pages={1--9},
  year={2026}
}

@article{deng2026taming,
  title={Taming OpenClaw: Security Analysis and Mitigation of Autonomous {LLM} Agent Threats},
  author={Deng, Xinhao and Zhang, Yixiang and Wu, Jiaqing and Bai, Jiaqi and Yi, Sibo and Zou, Zhuoheng and Xiao, Yue and Qiu, Rennai and Ma, Jianan and Chen, Jialuo and others},
  journal={arXiv preprint arXiv:2603.11619},
  pages={1--22},
  year={2026}
}

@article{cheng2026agent,
  title={Agent Privilege Separation in OpenClaw: A Structural Defense Against Prompt Injection},
  author={Cheng, Darren and Tsao, Wen-Kwang},
  journal={arXiv preprint arXiv:2603.13424},
  year={2026},
  pages={1--6}
}

@article{wang2026large,
  author={Wang, Yuntao and Pan, Yanghe and Su, Zhou and Deng, Yi and Zhao, Quan and Du, Linkang and Luan, Tom H. and Kang, Jiawen and Niyato, Dusit},
  journal={IEEE Communications Surveys \& Tutorials}, 
  title={Large Model-Based Agents: State-of-the-Art, Cooperation Paradigms, Security and Privacy, and Future Trends}, 
  year={2026},
  volume={28},
  number={},
  pages={1906-1949}
}

@article{dong2026clawdrain,
  title={Clawdrain: Exploiting tool-calling chains for stealthy token exhaustion in openclaw agents},
  author={Dong, Ben and Feng, Hui and Wang, Qian},
  journal={arXiv preprint arXiv:2603.00902},
  year={2026},
  pages={1--7}
}

@article{liu2026clawkeeper,
  title={{ClawKeeper}: Comprehensive Safety Protection for OpenClaw Agents Through Skills, Plugins, and Watchers},
  author={Liu, Songyang and Li, Chaozhuo and Wang, Chenxu and Hou, Jinyu and Chen, Zejian and Zhang, Litian and Liu, Zheng and Ye, Qiwei and Hei, Yiming and Zhang, Xi and others},
  journal={arXiv preprint arXiv:2603.24414},
  year={2026},
  pages={1--22}
}

@article{zhang2026clawworm,
  title={ClawWorm: Self-Propagating Attacks Across {LLM} Agent Ecosystems},
  author={Zhang, Yihao and Wei, Zeming and Luan, Xiaokun and Wu, Chengcan and Zhang, Zhixin and Wu, Jiangrong and Wu, Haolin and Chen, Huanran and Sun, Jun and Sun, Meng},
  journal={arXiv preprint arXiv:2603.15727},
  year={2026},
  pages={1--18}
}

@article{chen2026trajectory,
  title={A trajectory-based safety audit of clawdbot (openclaw)},
  author={Chen, Tianyu and Liu, Dongrui and Hu, Xia and Yu, Jingyi and Wang, Wenjie},
  journal={arXiv preprint arXiv:2602.14364},
  year={2026},
  pages={1--22}
}

@article{wang2026assistant,
  title={From assistant to double agent: Formalizing and benchmarking attacks on openclaw for personalized local {AI} agent},
  author={Wang, Yuhang and Xu, Feiming and Lin, Zheng and He, Guangyu and Huang, Yuzhe and Gao, Haichang and Niu, Zhenxing and Lian, Shiguo and Liu, Zhaoxiang},
  journal={arXiv preprint arXiv:2602.08412},
  year={2026},
  pages={1--11}
}

@article{chen2026openclaw,
  title={When OpenClaw {AI} Agents Teach Each Other: Peer Learning Patterns in the {Moltbook} Community},
  author={Chen, Eason and Guan, Ce and Elshafiey, Ahmed and Zhao, Zhonghao and Zekeri, Joshua and Shaibu, Afeez Edeifo and Prince, Emmanuel Osadebe},
  journal={arXiv preprint arXiv:2602.14477},
  year={2026},
  pages={1--7}
}

@article{li2026openclaw,
  title={OpenClaw {PRISM}: A Zero-Fork, Defense-in-Depth Runtime Security Layer for Tool-Augmented {LLM} Agents},
  author={Li, Frank},
  journal={arXiv preprint arXiv:2603.11853},
  year={2026},
  pages={1--23}
}

@article{deng2025ai,
  title={{AI} agents under threat: A survey of key security challenges and future pathways},
  author={Deng, Zehang and Guo, Yongjian and Han, Changzhou and Ma, Wanlun and Xiong, Junwu and Wen, Sheng and Xiang, Yang},
  journal={ACM Computing Surveys},
  volume={57},
  number={7},
  pages={1--36},
  year={2025}
}

@article{he2025emerged,
  title={The emerged security and privacy of {LLM} agent: A survey with case studies},
  author={He, Feng and Zhu, Tianqing and Ye, Dayong and Liu, Bo and Zhou, Wanlei and Yu, Philip S},
  journal={ACM Computing Surveys},
  volume={58},
  number={6},
  pages={1--36},
  year={2025}
}

@inproceedings{yu2025survey,
  title={A survey on trustworthy {LLM} agents: Threats and countermeasures},
  author={Yu, Miao and Meng, Fanci and Zhou, Xinyun and Wang, Shilong and Mao, Junyuan and Pan, Linsey and Chen, Tianlong and Wang, Kun and Li, Xinfeng and Zhang, Yongfeng and others},
  booktitle={Proceedings of ACM SIGKDD Conference on Knowledge Discovery and Data Mining (KDD)},
  pages={6216--6226},
  year={2025}
}

@ARTICLE{Wang2025IOA,
  author={Wang, Yuntao and Pan, Yanghe and Guo, Shaolong and Su, Zhou},
  journal={IEEE Open Journal of the Computer Society}, 
  title={Security of {Internet} of Agents: Attacks and Countermeasures}, 
  year={2025},
  volume={6},
  number={},
  pages={1611-1624},
  doi={10.1109/OJCS.2025.3589638}}

@ARTICLE{2026arXiv260216958D,
       author = {{Deng}, Xinhao and {Wu}, Jiaqing and {Chen}, Miao and {Xiao}, Yue and {Xu}, Ke and {Li}, Qi},
        title = "{Automating agent hijacking via structural template injection}",
      journal = {arXiv preprint 	arXiv:2602.16958},
pages={1--16},
         year = {2026},
}

@inproceedings{2026arXiv260105504D,
       author = {{Devarangadi Sunil}, Balachandra and {Sinha}, Isheeta and {Maheshwari}, Piyush and {Todmal}, Shantanu and {Mallik}, Shreyan and {Mishra}, Shuchi},
        title = {Memory poisoning attack and defense on memory based {LLM}-agents},
      booktitle = {Proceedings of Advances in Neural Information Processing Systems (NeurIPS)},
pages={1--19},
         year = {2025},
}

@ARTICLE{2025arXiv250217832H,
       author = {{Ha}, Hyeonjeong and {Zhan}, Qiusi and {Kim}, Jeonghwan and {Bralios}, Dimitrios and {Sanniboina}, Saikrishna and {Peng}, Nanyun and {Chang}, Kai-Wei and {Kang}, Daniel and {Ji}, Heng},
        title = "{MM-poisonRAG: Disrupting multimodal RAG with local and global poisoning attacks}",
      journal = {arXiv preprint 	arXiv:2502.17832},
pages={1--21},
         year = {2025},
}

@inproceedings{2025arXiv251023822Z,
       author = {{Zhang}, Zhenyu and {Chen}, Tianyi and {Xu}, Weiran and {Pentland}, Alex and {Pei}, Jiaxin},
        title = "{ReCAP: Recursive context-aware reasoning and planning for large language model agents}",
      booktitle = {Advances in Neural Information Processing Systems (NeurIPS)},
pages={1--29},
         year = {2025}
}

@article{CHOI201842,
title = {The role of privacy fatigue in online privacy behavior},
journal = {Computers in Human Behavior},
volume = {81},
pages = {42-51},
year = {2018},
issn = {0747-5632},
author = {Hanbyul Choi and Jonghwa Park and Yoonhyuk Jung},
}

@inproceedings{zhan2024injecagent,
  title         = {{InjecAgent}: Benchmarking Indirect Prompt Injections in Tool-Integrated Large Language Model Agents},
  author        = {Qiusi Zhan and Zhixiang Liang and Zifan Ying and Daniel Kang},
  booktitle       = {Findings of the Association for Computational Linguistics (ACL)},
  year          = {2024},
  pages={10471--10506}
}

@inproceedings{debenedetti2024agentdojo,
  author    = {Debenedetti, Edoardo and Zhang, Jie and Balunovic, Mislav and Beurer-Kellner, Luca and Fischer, Marc and Tram\`{e}r, Florian},
  title     = {{AgentDojo}: A Dynamic Environment to Evaluate Prompt Injection Attacks and Defenses for {LLM} Agents},
  booktitle = {Proceedings of Advances in Neural Information Processing Systems (NeurIPS)},
  volume    = {37},
  pages     = {82895--82920},
  year      = {2024}
}

@inproceedings{shi2026prompt,
  author    = {Shi, Jiawen and Yuan, Zenghui and Tie, Guiyao and Zhou, Pan and Gong, Neil Zhenqiang and Sun, Lichao},
  title     = {Prompt Injection Attack to Tool Selection in {LLM} Agents},
  booktitle = {Proceedings of the Network and Distributed System Security Symposium (NDSS)},
  year      = {2026},
  pages     = {1--18}
}

@inproceedings{zhang2025allies,
  title     = {From Allies to Adversaries: Manipulating {LLM} Tool-Calling through Adversarial Injection},
  author    = {Zhang, Rupeng and Wang, Haowei and Wang, Junjie and Li, Mingyang and Huang, Yuekai and Wang, Dandan and Wang, Qing},
  booktitle = {Proceedings of the Conference of the Nations of the Americas Chapter of the Association for Computational Linguistics: Human Language Technologies},
  pages     = {2009--2028},
  year      = {2025}
}

@inproceedings{hughes2024bon,
  title   = {Best-of-{N} Jailbreaking},
  author  = {Hughes, John and Price, Sara and Lynch, Aengus and Schaeffer, Rylan and Barez, Fazl and Koyejo, Sanmi and Sleight, Henry and Jones, Erik and Perez, Ethan and Sharma, Mrinank},
  booktitle    = {Advances in Neural Information Processing Systems (NeurIPS)},
  year    = {2025},
  pages     = {1--85}
}

@inproceedings{russinovich2025crescendo,
  title     = {Great, Now Write an Article About That: The Crescendo Multi-Turn {LLM} Jailbreak Attack},
  author    = {Russinovich, Mark and Salem, Ahmed and Eldan, Ronen},
  booktitle = {Proceedings of USENIX Security Symposium (USENIX Security)},
  year      = {2025},
pages     = {1--20}
}

@misc{mitre_t1567,
  author = {{MITRE ATT\&CK}},
  title  = {Exfiltration Over Web Service: T1567},
  year   = {2025},
  howpublished = {\url{https://attack.mitre.org/techniques/T1567/}},
  note   = {Version 1.5, last modified 24 October 2025}
}

@article{dhodapkar2026safetydrift,
  title         = {SafetyDrift: Predicting When {AI} Agents Cross the Line Before They Actually Do},
  author        = {Dhodapkar, Aditya and Pishori, Farhaan},
  journal       = {arXiv preprint arXiv:2603.27148},
  year          = {2026},
  pages     = {1--9}
}

@misc{mitreT1083,
  author = {{MITRE ATT\&CK}},
  title  = {{File and Directory Discovery}},
  year   = {2025},
  howpublished = {\url{https://attack.mitre.org/techniques/T1083/}},
  note   = {Technique T1083, Version 1.7, last modified October 24, 2025}
}

@misc{owasp2024llm06,
  author       = {{OWASP GenAI Security Project}},
  title        = {{LLM06}: Sensitive Information Disclosure},
  year         = {2024},
  howpublished = {\url{https://genai.owasp.org/llmrisk2023-24/llm06-sensitive-information-disclosure/}},
note = {Accessed on 2026-04-28}
}

@inproceedings{wang-etal-2025-unveiling-privacy,
  title     = {Unveiling Privacy Risks in {LLM} Agent Memory},
  author    = {Wang, Bo and He, Weiyi and Zeng, Shenglai and Xiang, Zhen and Xing, Yue and Tang, Jiliang and He, Pengfei},
  booktitle = {Proceedings of the Annual Meeting of the Association for Computational Linguistics (ACL)},
  pages     = {25241--25260},
  year      = {2025}
}

@article{alizadeh2025simple,
  title   = {Simple Prompt Injection Attacks Can Leak Personal Data Observed by {LLM} Agents During Task Execution},
  author  = {Alizadeh, Meysam and Samei, Zeynab and Stetsenko, Daria and Gilardi, Fabrizio},
  journal = {arXiv preprint arXiv:2506.01055},
  year    = {2025},
  pages     = {1--25}
}

@article{li2026secureagentskills,
  title   = {Towards Secure Agent Skills: Architecture, Threat Taxonomy, and Security Analysis},
  author  = {Li, Zhiyuan and Wu, Jingzheng and Ling, Xiang and Cui, Xing and Luo, Tianyue},
  journal = {arXiv preprint arXiv:2604.02837},
  year    = {2026},
  pages     = {1--27}
}

@misc{koi2026clawhavoc,
  author       = {{Koi Security}},
  title        = {ClawHavoc: 341 Malicious Clawed Skills Found by the Bot They Were Targeting},
  year         = {2026},
  month        = feb,
  howpublished = {\url{https://www.koi.ai/blog/clawhavoc-341-malicious-clawedbot-skills-found-by-the-bot-they-were-targeting}},
note = {Accessed on 2026-04-28}
}

@misc{beurerkellner2026toxicskills,
  author       = {Beurer-Kellner, Luca and Kudrinskii, Aleksei and Milanta, Marco and Nielsen, Kristian Bonde and Sarkar, Hemang and Tal, Liran},
  title        = {Snyk Finds Prompt Injection in 36\%, 1467 Malicious Payloads in a ToxicSkills Study of Agent Skills Supply Chain Compromise},
  year         = {2026},
  howpublished = {\url{https://snyk.io/blog/toxicskills-malicious-ai-agent-skills-clawhub/}},
note = {Accessed on 2026-02-18}
}

@article{tie2026badskill,
  title   = {{BadSkill}: Backdoor Attacks on Agent Skills via Model-in-Skill Poisoning},
  author  = {Tie, Guiyao and Shi, Jiawen and Zhou, Pan and Sun, Lichao},
  journal = {arXiv preprint arXiv:2604.09378},
  year    = {2026},
  pages     = {1--23}
}

@misc{oliveira2026amos,
  author       = {Oliveira, Alfredo and Tancio, Buddy and Fiser, David and Lin, Philippe and Reyes, Roel},
  title        = {Malicious {OpenClaw} Skills Used to Distribute Atomic macOS Stealer},
  year         = {2026},
  howpublished = {Trend Micro Research},
  howpublished = {\url{https://www.trendmicro.com/en_us/research/26/b/openclaw-skills-used-to-distribute-atomic-macos-stealer.html}},
note = {Accessed on 2026-02-18}
}

@misc{sgtech2026pinning,
  author       = {{Singapore Government Standards Portal}},
  title        = {Software Supply Chain},
  year         = {2026},
  howpublished = {\url{https://info.standards.tech.gov.sg/control-catalog/cybersecurity/sc/}},
  note         = {Includes SC-4 Dependency Manifest Version Pinning and SC-6 Dependency Installation during Deployment}
}

@misc{kydyraliev2024depconfusion,
  author       = {Kydyraliev, Meder},
  title        = {Defender's Perspective: Dependency Confusion and Typosquatting Attacks},
  year         = {2024},
  howpublished = {\url{https://slsa.dev/blog/2024/08/dep-confusion-and-typosquatting}},
note = {Accessed on 2026-01-20}
}

@misc{birsan2021dependencyconfusion,
  author       = {Birsan, Alex},
  title        = {Dependency Confusion: How {I} Hacked Into Apple, Microsoft and Dozens of Other Companies},
  year         = {2021},
  howpublished = {\url{https://medium.com/@alex.birsan/dependency-confusion-4a5d60fec610}},
note = {Accessed on 2026-01-20}
}

@article{liu2026agentskills,
  title   = {Agent Skills in the Wild: An Empirical Study of Security Vulnerabilities at Scale},
  author  = {Liu, Yi and Wang, Weizhe and Feng, Ruitao and Zhang, Yao and Xu, Guangquan and Deng, Gelei and Li, Yuekang and Zhang, Leo},
  journal = {arXiv preprint arXiv:2601.10338},
  year    = {2026},
  pages     = {1--23}
}

@article{shan2026openclaw,
  title   = {Don't Let the Claw Grip Your Hand: A Security Analysis and Defense Framework for OpenClaw},
  author  = {Shan, Zhengyang and Xin, Jiayun and Zhang, Yue and Xu, Minghui},
  journal = {arXiv preprint arXiv:2603.10387},
  year    = {2026},
  pages     = {1--12}
}

@misc{mitreG0139TeamTNT,
  author       = {{MITRE ATT\&CK}},
  title        = {{TeamTNT, Group G0139}},
  year         = {2025},
  howpublished = {MITRE ATT\&CK},
  url          = {https://attack.mitre.org/groups/G0139/},
  note         = {Accessed on 2026-05-06}
}

@misc{mitreT1105,
  author       = {{MITRE ATT\&CK}},
  title        = {Ingress Tool Transfer, Technique T1105},
  year         = {2025},
  howpublished = {MITRE ATT\&CK},
  note         = {Version 2.6, last modified October 24, 2025},
  howpublished = {\url{https://attack.mitre.org/techniques/T1105/}}
}

@misc{mitreT1027,
  author       = {{MITRE ATT\&CK}},
  title        = {{Obfuscated Files or Information, Technique T1027}},
  year         = {2025},
  howpublished = {\url{https://attack.mitre.org/techniques/T1027/}},
note = {Accessed on 2026-04-28}
}

@misc{quintero2026openclaw,
  author       = {Quintero, Bernardo},
  title        = {How {OpenClaw} {AI} Agent Skills Are Being Weaponized},
  year         = {2026},
  month        = feb,
  howpublished = {\url{https://blog.virustotal.com/2026/02/from-automation-to-infection-how.html}},
  note         = {VirusTotal Blog}
}

@misc{euler2023autogptRCE,
  title        = {Hacking {Auto-GPT} and Escaping Its Docker Container},
  author       = {Euler, Lukas},
  year         = {2023},
  howpublished = {\url {https://positive.security/blog/auto-gpt-rce}},
note = {Accessed on 2026-04-28}
}

@article{shi2025hafixagent,
  title         = {{HAFixAgent}: History-Aware Automated Program Repair Agent},
  author        = {Shi, Yu and Li, Hao and Adams, Bram and Hassan, Ahmed E.},
  journal       = {arXiv preprint arXiv:2511.01047},
  year          = {2025},
  pages     = {1--27}
}

@misc{owasp2025llm02,
  author       = {{OWASP Foundation}},
  title        = {{LLM02: Insecure Output Handling}},
  year         = {2025},
  howpublished = {\url{https://genai.owasp.org/llmrisk/llm02-insecure-output-handling/}},
note = {Accessed on 2026-04-28}
}

@misc{mitre2024cwe78,
  author       = {{MITRE Corporation}},
  title        = {{CWE-78: Improper Neutralization of Special Elements used in an OS Command ('OS Command Injection')}},
  year         = {2024},
  howpublished = {\url{https://cwe.mitre.org/data/definitions/78.html}},
note = {Accessed on 2026-04-28}
}

@misc{irwin2025code,
  author       = {John Irwin and Kai Greshake},
  title        = {How Code Execution Drives Key Risks in Agentic {AI} Systems},
  year         = {2025},
  month        = nov,
  organization = {NVIDIA Developer Blog},
  howpublished = {\url{https://developer.nvidia.com/blog/how-code-execution-drives-key-risks-in-agentic-ai-systems/}},
note = {Accessed on 2026-04-16}
}

@article{wang2026agentassetrealworldsafety,
  title         = {Your Agent, Their Asset: A Real-World Safety Analysis of OpenClaw},
  author        = {Zijun Wang and Haoqin Tu and Letian Zhang and Hardy Chen and Juncheng Wu and Xiangyan Liu and Zhenlong Yuan and Tianyu Pang and Michael Qizhe Shieh and Fengze Liu and Zeyu Zheng and Huaxiu Yao and Yuyin Zhou and Cihang Xie},
  journal       = {arXiv preprint arXiv:2604.04759},
  year          = {2026},
  pages     = {1--19}
}

@misc{bors2026escaping,
  author       = {David Bors},
  title        = {Escaping the Agent: On Ways to Bypass OpenClaw's Security Sandbox},
  year         = {2026},
  month        = feb,
  organization = {Snyk Labs},
  howpublished = {\url{https://labs.snyk.io/resources/bypass-openclaw-security-sandbox/}},
note = {Accessed on 2026-04-14}
}

@article{banerjee2026cascade,
  title         = {Cascade: Composing Software-Hardware Attack Gadgets for Adversarial Threat Amplification in Compound {AI} Systems},
  author        = {Banerjee, Sarbartha and Sahu, Prateek and Vahldiek-Oberwagner, Anjo and Sanchez Vicarte, Jose and Tiwari, Mohit},
  journal       = {arXiv preprint arXiv:2603.12023},
  year          = {2026},
  pages     = {1--11}
}

@article{bousetouane2026memorycontrol,
  title   = {{AI} Agents Need Memory Control Over More Context},
  author  = {Bousetouane, Fouad},
  journal = {arXiv preprint arXiv:2601.11653},
  year    = {2026},
  pages={1--32}
}

@inproceedings{yang2026zombie,
  title   = {Zombie Agents: Persistent Control of Self-Evolving {LLM} Agents via Self-Reinforcing Injections},
  author  = {Yang, Xianglin and He, Yufei and Ji, Shuo and Hooi, Bryan and Dong, Jin Song},
  booktitle = {ICLR 2026 Workshop on Lifelong Agents: Learning, Aligning, Evolving},
  year    = {2026},
  pages     = {1--14}
}

@article{shi2025progent,
  title         = {Progent: Programmable Privilege Control for {LLM} Agents},
  author        = {Shi, Tianneng and He, Jingxuan and Wang, Zhun and Li, Hongwei and Wu, Linyu and Guo, Wenbo and Song, Dawn},
  journal       = {arXiv preprint arXiv:2504.11703},
  year          = {2025},
  pages     = {1--30}
}

@article{shapira2026agentschaos,
  title         = {Agents of Chaos},
  author        = {Shapira, Natalie and Wendler, Chris and Yen, Avery and others},
  year          = {2026},
  journal       = {arXiv preprint arXiv:2602.20021},
  pages     = {1--84}
}

@misc{owasp2025unbounded,
  author = {{OWASP GenAI Security Project}},
  title  = {{LLM10:2025 Unbounded Consumption}},
  year   = {2025},
  howpublished = {\url{https://genai.owasp.org/llmrisk/llm102025-unbounded-consumption/}},
note = {Accessed on 2026-04-18}
}

@article{mahara2025oauth,
      title={Detecting Malicious Entra {OAuth} {Apps} with {LLM}-Based Permission Risk Scoring}, 
      author={Ashim Mahara},
      year={2025},
      journal       = {arXiv preprint arXiv:2512.15781},
      pages={1--54} 
}

@article{zhao2026clawguard,
      title={{ClawGuard}: A Runtime Security Framework for Tool-Augmented {LLM} Agents Against Indirect Prompt Injection}, 
      author={Wei Zhao and Zhe Li and Peixin Zhang and Jun Sun},
      year={2026},
      journal       = {arXiv preprint arXiv:2604.11790},
      pages={1--19} 
}

@article{wen2026agentsys,
      title={{AgentSys}: Secure and Dynamic {LLM} Agents Through Explicit Hierarchical Memory Management}, 
      author={Ruoyao Wen and Hao Li and Chaowei Xiao and Ning Zhang},
      year={2026},
journal       = {arXiv preprint arXiv:2602.07398},
      pages={1--21}
}

@article{wei2025memguard,
  title={{A-MemGuard}: A proactive defense framework for {LLM}-based agent memory},
  author={Wei, Qianshan and Yang, Tengchao and Wang, Yaochen and Li, Xinfeng and Li, Lijun and Yin, Zhenfei and Zhan, Yi and Holz, Thorsten and Lin, Zhiqiang and Wang, XiaoFeng},
  journal={arXiv preprint arXiv:2510.02373},
  year={2025},
      pages={1--27}
}

@articles{AgentDyn,
  title={{AgentDyn}: Are Your Agent Security Defenses Deployable in Real-World Dynamic Environments?},
  author={Hao Li and Ruoyao Wen and Shanghao Shi and Ning Zhang and Yevgeniy Vorobeychik and Chaowei Xiao},
  journal={arXiv preprint arXiv:2602.03117},
  year={2026},
      pages={1--23}
}

@article{hu2026maltool,
  title={{MalTool}: Malicious tool attacks on {LLM} agents},
  author={Hu, Yuepeng and Jia, Yuqi and Li, Mengyuan and Song, Dawn and Gong, Neil},
  journal={arXiv preprint arXiv:2602.12194},
  year={2026},
      pages={1--34}
}

@article{li2025stac,
  title={{STAC}: When Innocent Tools Form Dangerous Chains to Jailbreak {LLM} Agents},
  author={Li, Jing-Jing and He, Jianfeng and Shang, Chao and Kulshreshtha, Devang and Xian, Xun and Zhang, Yi and Su, Hang and Swamy, Sandesh and Qi, Yanjun},
  journal={arXiv preprint arXiv:2509.25624},
  year={2025},
      pages={1--30}
}

@article{zhao2025mind,
  title={Parasites in the Toolchain: A Large-Scale Analysis of Attacks on the {MCP} Ecosystem},
  author={Zhao, Shuli and Hou, Qinsheng and Zhan, Zihan and Wang, Yanhao and Xie, Yuchong and Guo, Yu and Chen, Libo and Li, Shenghong and Xue, Zhi},
  booktitle = {Proceedings of IEEE Symposium on Security and Privacy (S&P)},
  year={2026},
  pages     = {1--18}
}

@article{10.1145/3704724,
author = {Sood, Aditya K. and Zeadally, Sherali},
title = {Malicious {AI} Models Undermine Software Supply-Chain Security},
year = {2025},
volume = {68},
number = {6},
journal = {Communications of the ACM},
pages = {62--71}
}

\end{document}